\documentclass[10pt,twocolumn,letterpaper]{article}

\usepackage[pagenumbers]{iccv} % To force page numbers, e.g. for an arXiv version
\usepackage{overpic}

\definecolor{iccvblue}{rgb}{0.21,0.49,0.74}
\usepackage[pagebackref,breaklinks,colorlinks,allcolors=iccvblue]{hyperref}

\def\confName{ICCV}
\def\confYear{2025}

\title{MMGen: Unified Multi-modal Image Generation and \\  Understanding in One Go}

\author{{Jiepeng Wang$^{1,3,\ast}$, 
Zhaoqing Wang$^{2,3,\ast}$,
Hao Pan$^{4}$,
Yuan Liu$^{5}$,
Dongdong Yu$^{3}$,} \\
{Changhu Wang$^{3,\dagger}$,
Wenping Wang$^{6}$} \\
$^1$The University of Hong Kong, $^2$ The University of Sydney, $^3$AIsphere, $^4$Tsinghua University, \\
$^5$Hong Kong University of Science and Technology,  $^6$Texas A\&M University
}

\begin{document}
\twocolumn[{%
        \vspace{-18pt}
	\maketitle
	\renewcommand\twocolumn[1][]{#1}%
	\begin{center}
		\centering
            \begin{overpic}[width=0.95\textwidth]{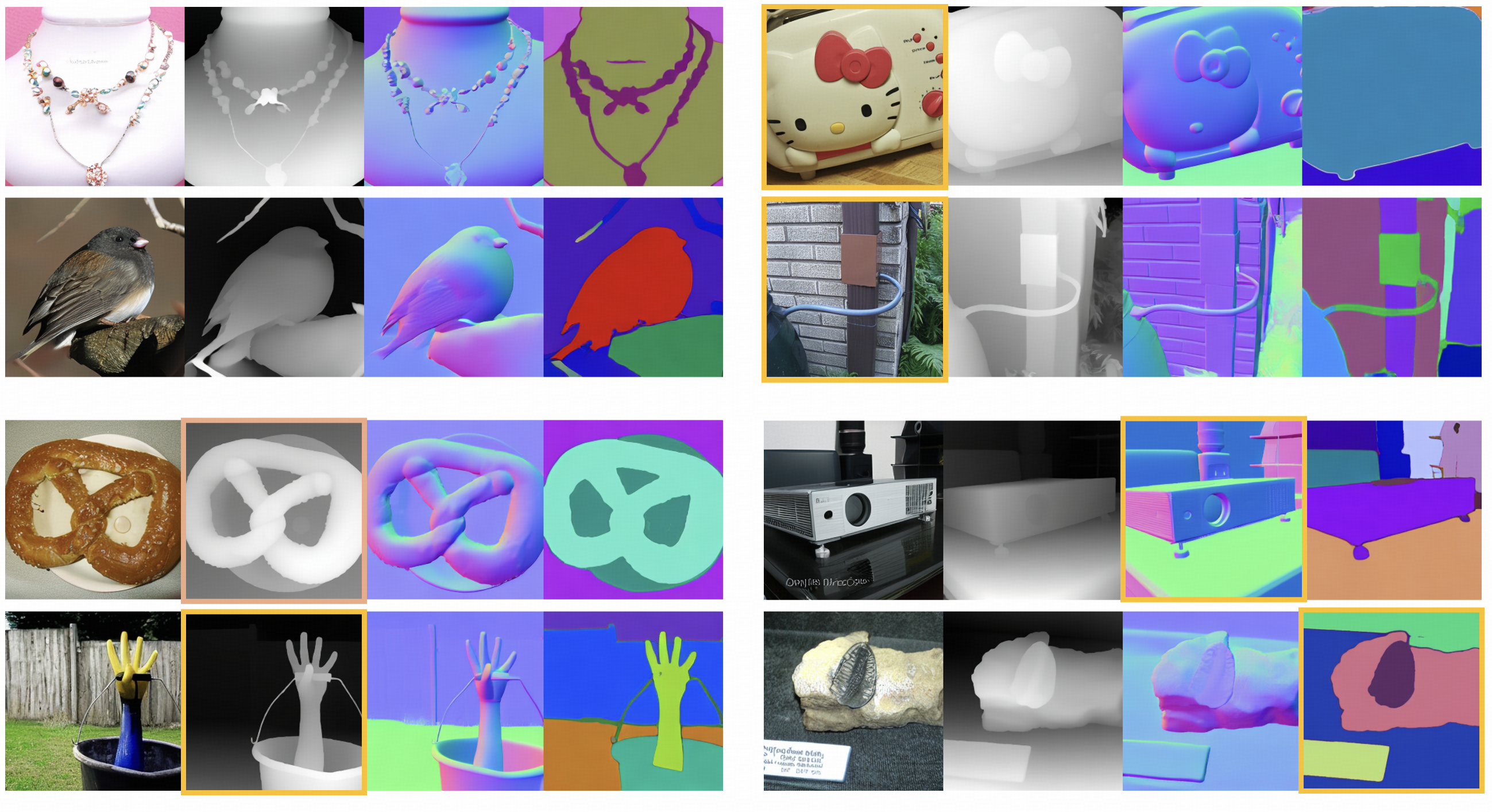}
            \put(4,55){\small rgb}
            \put(16.5,55){\small depth}
            \put(27.5,55){\small normal}
            \put(37.5,55){\small segmentation}

            \put(55,55){\small rgb}
            \put(67.5,55){\small depth}
            \put(77.5,55){\small normal}
            \put(87.5,55){\small segmentation}

            \put(6,27){\small (a) Multi-modal category-conditioned generation}
            \put(61,27){\small (b) Multi-modal visual understanding}
            \put(34,-1.0){\small (c) Multi-modal conditioned generation}
          \end{overpic}
        \captionof{figure}{\textbf{Unified multi-modal generation and understanding in a single diffusion process.} We present a unified framework, capable of handling multi-modal generation and understanding in one model: (a) \textbf{Multi-modal category-conditioned generation}: Given the category information, multi-modal images (i.e., rgb, depth, normal, semantic segmentation) are generated simultaneously in a single diffusion process; (b) \textbf{Multi-modal visual understanding}: Given a reference image (highlighted with yellow rectangles), our framework accurately estimates the associated depth, normal, and semantic segmentation results; (c) \textbf{Multi-modal conditioned generation}: Given a fine-grained condition input (e.g., depth or normal, highlighted by yellow rectangles), our model can accurately generate the corresponding rgb image and other aligned outputs in parallel. Each row illustrates one example per condition. 
         } 
		\label{fig:teaser}
	\end{center}
}
]

\renewcommand{\thefootnote}{$\ast$}
\footnotetext[1]{Denotes equal contribution}
\renewcommand{\thefootnote}{$\dagger$}
\footnotetext[2]{Denotes corresponding author}

\begin{abstract}
A unified diffusion framework for multi-modal generation and understanding has the transformative potential to achieve seamless and controllable image diffusion and other cross-modal tasks. 
In this paper, we introduce MMGen, a unified framework that integrates multiple generative tasks into a single diffusion model. This includes: (1) multi-modal category-conditioned generation, where multi-modal outputs are generated simultaneously through a single inference process, given category information; (2) multi-modal visual understanding, which accurately predicts depth, surface normals, and segmentation maps from RGB images; and (3) multi-modal conditioned generation, which produces corresponding RGB images based on specific modality conditions and other aligned modalities. Our approach develops a novel diffusion transformer that flexibly supports multi-modal output, along with a simple modality-decoupling strategy to unify various tasks. Extensive experiments and applications demonstrate the effectiveness and superiority of MMGen across diverse tasks and conditions, highlighting its potential for applications that require simultaneous generation and understanding.
Our project page: \url{https://jiepengwang.github.io/MMGen/}.

\end{abstract}  
\section{Introduction}
\label{sec:intro}

Humans possess an exceptional ability to perceive and imagine information of visual scenes in a multi-modal manner \cite{tang2025humanfoundationmodel}. 
When we imagine/look at a scene, we can mentally construct the composition of objects, their spatial relationships, and aspects of geometry like depth and normals.
This capacity of multi-modal imagination enables us to anticipate scenarios and simulate possible future complex interactions.
Emulating this human-like, multi-faceted capacity for both perceiving and imagining in artificial intelligence systems is significant for downstream applications. 

To this end, advancements in diffusion-based image generation techniques have opened new possibilities, with recent models demonstrating impressive performance in producing high-quality and diverse RGB images \cite{ho2020ddpm, rombach2022latentdiff,peebles2023dit,ma2024sit}.
And to achieve conditional control, many methods have introduced fine-tuning techniques to incorporate various conditional inputs, such as bounding boxes, depth, normal maps, and layout guidance \cite{xie2023boxdiff, zheng2023layoutdiffusion, yang2023lawdiff, mou2024t2i_adapter, zhang2023controlnet, li2023gligen, 4m}. Additionally, several approaches leverage large-scale depth and normal data to enhance diffusion models' capabilities for visual understanding \cite{bochkovskii2024depthpro, ye2024stablenormal, he2024lotus,fu2025geowizard}. These methods collectively show that generation and visual understanding capabilities are inherently achievable within large-scale diffusion models.
However, most existing models primarily focus on excelling in a single task—either generation or visual understanding. 
Consequently, for downstream tasks requiring multi-modal information, we often need to run different large-scale foundation models separately, which is computationally intensive and time-consuming. 
For instance, in depth-conditioned image generation, like ControlNet \cite{zhang2023controlnet}, a dedicated depth estimation model is first needed to extract depth information from the reference image before using it as a condition for generation.
Therefore, incorporating visual understanding capabilities into a generative model is a promising direction to enable more efficient, flexible and comprehensive multi-modal tasks, such as ControlNet-like generation and 3D reconstruction (Refer to Fig. \ref{fig:3drecon}).

Recent efforts have aimed to unify multi-modal capabilities within one diffusion process \cite{krishnan2025orchid,wang2024diffx}. For instance, DiffX \cite{wang2024diffx} proposes a Multi-Path Variational AutoEncoder (VAE) \cite{kingma2013vae} to encode various visual modalities, such as RGB and depth, into a single shared latent space, enabling diffusion across on it. By employing separate decoders for each modality, DiffX can produce modality-specific outputs from this joint latent representation, allowing for synchronized cross-modal synthesis. However, DiffX and similar models \cite{krishnan2025orchid}, are constrained by tightly coupled modalities, which limits their flexibility and scalability. 
In this context, "coupled modality" refers to the fact that multiple modalities are jointly encoded into a shared latent space via VAE before diffusion occurs. As a result, it is not possible to use one modality as a condition to generate the others independently in the diffusion process. 
Addressing these limitations through a modality-decoupling strategy could provide independent control over each modality as condition signals within a unified framework, 
enhancing flexibility in multi-modal generation and understanding.

To bridge this gap, we introduce MMGen, a novel framework designed to emulate the human-like capacity for both multi-modal image generation and visual understanding within a single diffusion model, more importantly in one diffusion process.  In this paper, we focus on 4 representative visual modalities: RGB, depth, normal and segmentation.
Specifically, we utilize a pretrained Variational Autoencoder (VAE) \cite{rombach2022latentdiff} to encode each modality into latent patch representations, ensuring consistent encoding quality across modalities. Building upon the SiT architecture \cite{ma2024sit}, the encoded multi-modal patches corresponding to the same image location are grouped to form the multi-modal patch input, which is blended with random noise to initiate the diffusion process. Our novel MM Diffusion model, designed to support both multi-modal inputs and outputs, employs modality-specific decoding heads, enabling each modality's unique attributes to be preserved during generation. To further decouple modalities, we introduce a modality-decoupling strategy with distinct denoising schedules for each modality and learnable task embeddings to enhance modality decoupling. 
Finally, the denoised patches are reprojected to their original spatial locations for each modality and decoded back into image pixels, providing complete, high-quality outputs for each modality.

Building upon our MMGen framework, a comprehensive range of tasks can be supported within one diffusion process.
The key capabilities of our approach enable the following applications: (1) Multi-modal category-conditioned generation: By leveraging a single diffusion process, our framework can generate diverse, multi-modal images simultaneously, conditioned on specified categories. 
This allows MMGen to capture and represent a wide range of scene attributes within a unified process. 
(2) Multi-modal conditioned generation: MMGen also supports generation based on specific conditions, such as depth maps, normals, or masks. 
This process allows the generation of both RGB and the other synchronized, modality-aligned outputs, which are essential for applications requiring precise cross-modal synthesis and control.
(3) Multi-modal visual understanding: Our framework can accurately estimate multiple scene properties simultaneously, including depth, surface normals, and semantic segmentation, for the input images. 
This capability enhances interpretability and utility in analytical tasks, making MMGen versatile for applications requiring detailed scene comprehension. 

We train and evaluate our method’s generation performance, including both category-conditioned and conditioned generation, on the ImageNet-1k dataset \cite{deng2009imagenet}. 
To quantitatively assess visual understanding performance, we test our method on the widely used ScanNet dataset \cite{dai2017scannet}. 
We adopt the same architecture as SiT \cite{ma2024sit} with similar model parameters.
Experiments show that our method achieves comparable generation performance of SiT while extending its capabilities to support category-conditioned generation with multi-modal outputs, fine-grained conditioned generation, and multi-modal visual understanding.
These results demonstrate the flexibility and coherence of our model and highlight its potential for real-world applications where simultaneous generation and understanding are essential.

\section{Related Works}
\label{sec:rw}

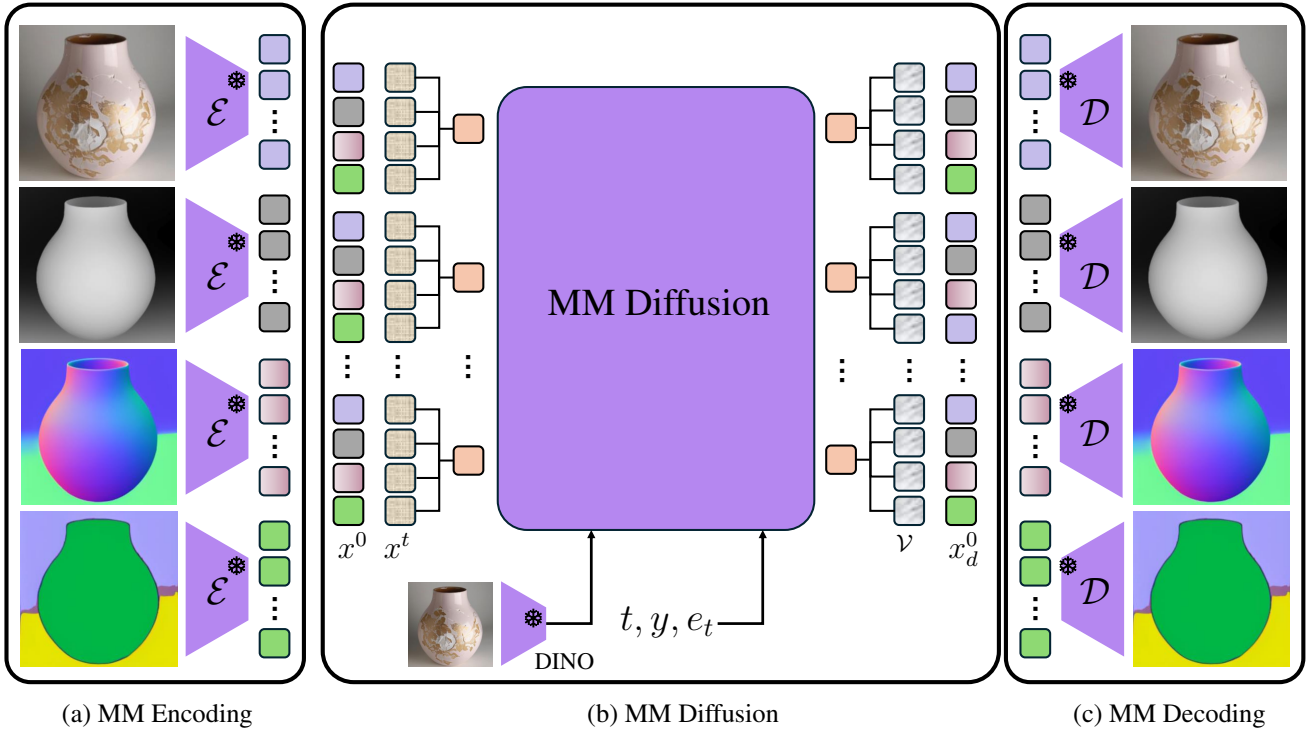
\begin{figure*}
    \centering
    \begin{overpic}[width=\linewidth]{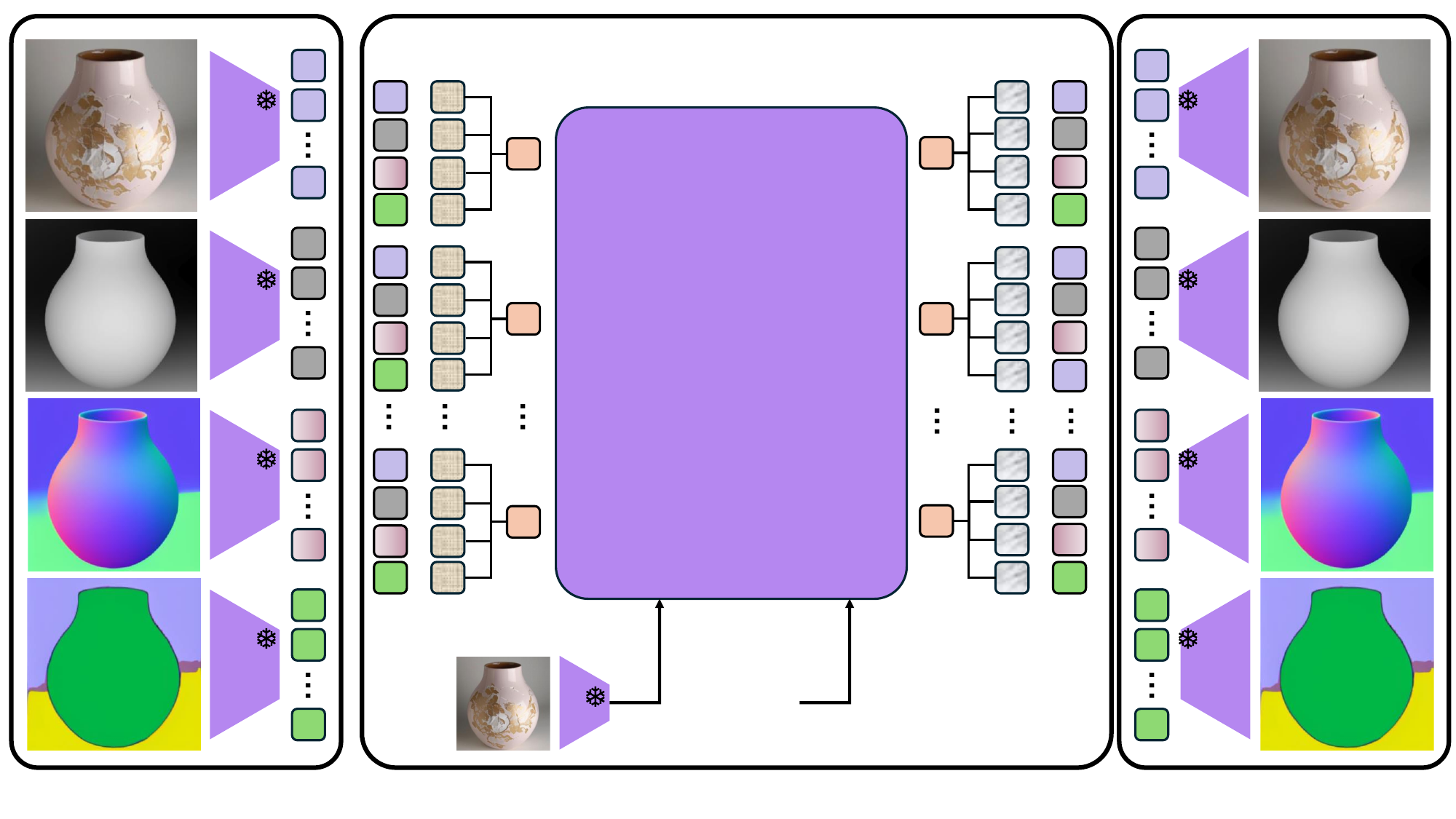}
        \put(5,-1){(a) MM Encoding}
        \put(45,-1){(b) MM Diffusion}
        \put(82,-1){(c) MM Decoding}

        \put(16,44.5){\Large $\mathcal{E}$}
        \put(16,32){\Large $\mathcal{E}$}
        \put(16,20){\Large $\mathcal{E}$}
        \put(16,8){\Large $\mathcal{E}$}

        \put(82.5,44.5){\Large $\mathcal{D}$}
        \put(82.5,32){\Large $\mathcal{D}$}
        \put(82.5,20){\Large $\mathcal{D}$}
        \put(82.5,8){\Large $\mathcal{D}$}

        \put(41,3){\small DINO}
        \put(47.5,6){\Large $t, y, e_t$}

        \put(42,30){\Large MM Diffusion}

        \put(26,   11.5){\large  ${x}^0 $}
        \put(29, 11.5){ \large  $x^t $}

       \put(72.5,   11.5){\large  ${x}^0_d $}
        \put(68, 12){ \small  $\mathcal{V} $}
        
    \end{overpic}
    \caption{\textbf{Method overview.} (1) MM Encoding: Given paired multi-modal images, we first use a shared pretrained VAE encoder to encode each modality into latent patch codes. (2) MM Diffusion: Patch codes corresponding to the same image location are grouped to form the multi-modal patch input \( x^0 \), which is blended with random noise to create the diffusion input \( x_t \). Conditioned on timestep \( y \), category label \( t \) and task embedding \( e_t \), the MM Diffusion model iteratively predicts the velocity, resulting in denoised multi-modal patches \( x^0_d \). (3) MM Decoding: Finally, these patches are reprojected to the original image locations for each modality and decoded back into image pixels using a shared pretrained VAE decoder.
    }
    \vspace{-4pt}
    \label{fig:method_overview}
\end{figure*}

\paragraph{Controllable image diffusion}  
Large diffusion models (LDMs) have shown impressive capabilities in generating high-quality, diverse images \cite{ho2020ddpm, rombach2022latentdiff, ramesh2022hierarchicaldiff, peebles2023dit, ma2024sit, yu2024repa, nichol2021improvedddpm, nichol2021glide}, often pretrained on large-scale datasets \cite{schuhmann2022laion, deng2009imagenet}. 
Building on significant progress in large-scale text-to-image generation models, 
many works explore empower the diffusion models with
the ability to (1) use reference images and other conditional inputs, such as depth or normal maps, to control the image generation process \cite{xie2023boxdiff,zheng2023layoutdiffusion,yang2023lawdiff,mou2024t2i_adapter,zhang2023controlnet,li2023gligen,4m} and (2) precisely localize concepts and understand visual contents \cite{he2024lotus,ye2024stablenormal,yang2024depthanyvideo,ke2023marigold,fu2025geowizard},  such as depth, normal and segmentation. For controllable image diffusion, a notable advancement is ControlNet \cite{zhang2023controlnet}, which enables controllable generation by fine-tuning a pretrained text-to-image diffusion model with various conditional inputs. 
For generative visual understanding, these methods usually finetune a pre-trained diffusion model to adapt a new visual modality, like Marigold \cite{ke2023marigold} for depth estimation.

While these diffusion-based methods have advanced single-modality-based generation or understanding, 
they typically need to be fine-tuned for each modality or are often restricted to generate only RGB images, lacking the flexibility to handle simultaneous multi-modal outputs. 
MMGen addresses this limitation by unifying multiple visual signals (depth, normals, and segmentation) within a single diffusion model, allowing simultaneous multi-modal understanding and generation, without requiring additional fine-tuning for each condition.

\paragraph{Unified multi-modal image diffusion and understanding}
Recently, several concurrent works have attempted to unify various generation and understanding tasks within a single diffusion framework \cite{le2024onediff,chen2024UniReal}. OneDiff\cite{le2024onediff}, for instance, treats different image-level tasks as a sequence of image views with varying noise scales during training, enabling both image generation and understanding within a single model. 
Additionally, many non-diffusion-based approaches \cite{lu2022unifiediov1,lu2024unifiediov2,bachmann2022multimae,khirodkar2024sapiens,4m,bachmann20254m21}, explore unifying multiple modalities into one model. However, these methods either generate only one modality per inference or treat multiple modalities as different image views, both of which lead to higher computational costs. Meanwhile, other methods focus exclusively on generation tasks \cite{zhu2024consistent,bao2022generativemodeling}. 
In contrast, our approach enables not only multi-modal generation in a unified model but also in one single diffusion process.
Rather than treating different modalities as a sequence of image views, our method significantly reduces computational overhead while maintaining a cost comparable to pure RGB generation. 

Despite these advancements, few works focus on generating multi-modal images simultaneously in a single diffusion process \cite{wang2024diffx,krishnan2025orchid}. DiffX \cite{wang2024diffx} introduces a Multi-Path Variational AutoEncoder (VAE) \cite{kingma2013vae} to encode different modalities into a shared latent space, enabling diffusion on this latent representation. Through multi-path decoders, DiffX decodes the denoised latent results back into individual modalities, achieving high-quality, cross-modal synthesis. Similarly, MT-Diffusion \cite{chen2024mtdiffusion} proposes a multi-task loss to generate multi-modalities and adopts learnable heads to decode each multi-modality.
However, these methods tightly couples modalities, limiting flexibility and scalability. In contrast, MMGen’s modality-decoupling strategy allows independent control over each modality within a unified framework, supporting diverse modality combinations. 
\section{Method}

In this section, we introduce MMGen, a unified framework for multi-modal generation and understanding. 
This section is organized into three parts: (1) Preliminary (Sec. \ref{sec:pre}), covering foundational principles of diffusion; 
(2) Multi-modal generation (Sec. \ref{sec:mmgen}), describing MMGen’s design; and (3) Training (Sec. \ref{sec:train}), outlining the loss functions for optimization. 
Note that to train MMGen, we first prepared an aligned multi-modal dataset via 2D foundation models, 
denoted as $ \mathcal{I}_{mm} = \{ ( \mathcal{I}_r, \mathcal{I}_d, \mathcal{I}_n, \mathcal{I}_s ) \mid \mathcal{I}_r \in \text{ImageNet-1k}  \} $,
including aligned RGB, depth, normal and segmentation. Please refer to the supplementary material for more details.
Together, these components form a cohesive framework that supports flexible and effective multi-modal generation and understanding. Fig. \ref{fig:method_overview} shows an overview of MMGen.

\subsection{Preliminary}\label{sec:pre}
SiT \cite{ma2024sit} is a flow and diffusion-based framework that models data generation as a continuous transformation between data and noise. In SiT, the forward process gradually adds noise to the data, creating a smooth path from the original data distribution to pure noise, which is then reversed during generation. The forward process is defined by blending the original data \( \mathbf{x}_0 \) with Gaussian noise \( \epsilon \sim \mathcal{N}(0, \mathbf{I}) \), forming a latent variable \( \mathbf{x}_t \) at each time step \( t \in [0, 1] \):

\[
\mathbf{x}^t = t \cdot \mathbf{x}^0 + (1 - t) \cdot \epsilon
\]

This process can be represented by a probability flow ordinary differential equation (PF ODE), which models the evolution of \( \mathbf{x}_t \) over time through a velocity field \( \mathbf{v}(\mathbf{x}_t, t) = d\mathbf{x}_t / d\mathbf{t}\).
To learn this velocity field, a neural network \( \mathbf{v}_\theta(\mathbf{x}_t, t) \) is trained to approximate the target velocity \( \mathbf{v}^* = \mathbf{x}_0 - \epsilon \).
The network is optimized by minimizing the velocity loss \( \mathcal{L}_{\text{velocity}} \), defined as:
\[
\mathcal{L}_{\text{velocity}}(\theta) := \mathbb{E}_{\mathbf{x}^0, \epsilon, t} \left[ \left\| \mathbf{v}_\theta(\mathbf{x}^t, t) - \mathbf{v}^* \right\|^2 \right]
\]

\subsection{MMGen}\label{sec:mmgen}
Given the aligned dataset \( \mathcal{I}_{\text{mm}} \), MMGen is trained to perform both multi-modal generation and visual understanding within a unified framework. The architecture consists of two main components: 1) MM encoding and decoding (Fig. \ref{fig:method_overview} (a) and (c)) and 2) MM diffusion (Fig. \ref{fig:method_overview} (b)).

\paragraph{MM encoding and decoding}  
The MM encoding and decoding component is responsible for transforming multi-modal inputs into a shared latent space and reconstructing them back into their respective modalities. Given paired multi-modal images from the aligned dataset \( \mathcal{I}_{\text{mm}} = \{ \mathcal{I}_r, \mathcal{I}_d, \mathcal{I}_n, \mathcal{I}_s \} \), we use a shared pretrained Variational Autoencoder (VAE) \cite{rombach2022latentdiff} encoder to encode each modality in  \(\mathcal{I}_\text{mm} \)
into latent representations
\(\mathcal{X}_{\text{mm}} = \{ \mathbf{x}^{0}_{r}, \mathbf{x}^{0}_{d}, \mathbf{x}^{0}_{n}, \mathbf{x}^{0}_{s} \} \). Here, \( \mathbf{x}^{0}_{r} \), \( \mathbf{x}^{0}_{d} \), \( \mathbf{x}^{0}_{n} \), and \( \mathbf{x}^{0}_{s} \) represent the encoded latent tokens for RGB, depth, normal, and segmentation, respectively.

After processing through the MM Diffusion model, these denoised multi-modal patches \( \mathcal{X}_{\text{mm}}^d \) are reprojected to their original spatial configurations for each modality. A shared VAE decoder then reconstructs each modality's output from the denoised latents back into original image forms.

\paragraph{MM diffusion}
The MM diffusion component is the core of MMGen’s multi-modal processing, leveraging a diffusion process inspired by SiT to iteratively denoise multi-modal latent representations. This component enables MMGen to synthesize aligned outputs across RGB, depth, normal, and segmentation modalities in a unified manner.

Starting from the multi-modal latent representations \( \mathcal{X}_{\text{mm}} = \{ \mathbf{x}^{0}_{r}, \mathbf{x}^{0}_{d}, \mathbf{x}^{0}_{n}, \mathbf{x}^{0}_{s} \} \) obtained from MM Encoding, we group the latent patches corresponding to the same spatial location across modalities. Let \( \mathcal{X}_{\text{mm}}^g = \{ \mathbf{x}^{0, i}_{\text{mm}} \mid i = 1, \dots, n \} \) represent the grouped multi-modal patches, where \( i \) indexes each spatial location (or patch) and \( \mathbf{x}^{0, i}_{\text{mm}} = (\mathbf{x}^{0, i}_{r}, \mathbf{x}^{0, i}_{d}, \mathbf{x}^{0, i}_{n}, \mathbf{x}^{0, i}_{s}) \) denotes the multi-modal latent codes at the \( i \)-th location. The leftmost column in Fig. \ref{fig:method_overview} (b) shows a visualization of grouped patches.

For these grouped patches, each modality is first blended with random noise to produce the noisy input at time \( t_m \) (\( m \in \mathcal{M}= \{ r, d, n, s \} \)), respectively:

\[
\mathbf{x}_m^t = t_m \cdot \mathbf{x}_m^0 + (1 - t_m) \cdot \epsilon_m
\]

Then these blended patches of all modalities will be fused via Multi-layer Perceptrons (MLPs) into a single latent vector as the input to the MM Diffusion Transformer,
During the iterative denoising process, the output of MM Diffusion Transformer predicts the velocity field \( \mathbf{v}_\theta(\mathbf{x}^t_m, t_m) \) for each modality \( m \) via different learnable decoding head, guiding \( \mathbf{x}_m^t \) back toward the clean, denoised multi-modal patch \( \mathcal{X}_{mm}^d\). 
Please refer to Sec. 1.3 and Fig. 1 in the supplementary for more discussions of our design.

\paragraph{Modality decoupling}
A distinctive feature of MM Diffusion is its modality-decoupling strategy, which assigns separate denoising schedules to each modality. By allowing each modality to follow its own independent denoising schedule, the model can adjust each modality independently while maintaining coherence across them. To achieve this, the time embeddings of different modalities \( t_m \), denoted as \( t_r \), \( t_d \), \( t_n \), and \( t_s \) for RGB, depth, normal, and segmentation respectively, are fused into a single fused time embedding \( t_{\text{fused}} \) through a multi-layer perceptron (MLP):

\[
t_{\text{fused}} = \text{MLP}(t_r, t_d, t_n, t_s)
\]

This modality-decoupling strategy enhances flexibility, enabling applications such as category-conditioned generation, conditioned generation, and visual understanding by allowing the model to selectively control each modality in a coordinated yet independent manner.

In preliminary experiments, we found that using only the different time-embedding strategies was insufficient for the model to fully differentiate between tasks. To improve the model's capacity for handling diverse tasks, we introduce additional task embedding tokens \( e_t \) as part of the model’s conditioning input. Specifically, we use learnable tokens to represent different tasks, including category-conditioned generation, conditioned generation and visual understanding, allowing the model to better distinguish between them. These task embeddings are combined with the time embedding \( t \) and category label embedding \( y \), creating a unified conditioning input:

\[
c = f_c(e_t, t_{\text{fused}}, y)
\]

where \( f_c(\cdot) \) denotes the fusion function, i.e., addition. 

This combined conditioning \( c \) is then injected into the model, enabling it to leverage explicit task information alongside time and category embeddings, thereby enhancing its ability to handle various tasks.

\subsection{Training}\label{sec:train}
The training of MMGen incorporates two main losses: a velocity loss with random modality drop augmentation and a representation alignment regularization. These losses guide the model to learn effective multi-modal representations and improve its flexibility across various tasks.

\paragraph{Velocity loss}  
The primary training objective in MM diffusion is the velocity loss, which encourages accurate prediction of the target velocity for each modality. To enhance the model’s robustness across different modality combinations, we apply random modality drop augmentation during training, where the supervision of one or more modalities (except RGB) is randomly dropped in each iteration. Our findings indicate that RGB is the most challenging modality, and this strategy helps the model focus more on RGB while adapting to partial information, promoting flexibility across various tasks. The velocity loss is formulated as:

\[
\mathcal{L}_{\text{v}} := \sum_{m \in \mathcal{M}} \mathbb{E}_{\mathbf{x}^{0}_m, \epsilon_m, t_m} \left[ \left\| \mathbf{v}_\theta(\mathbf{x}_t^m, t_m) - \mathbf{v}^{*}_m \right\|^2  \right] \cdot \mathbf{1}_{ \{p>0.5\}   }
\]

where 
\( \mathbf{1}_{\{p>0.5\}} \) is an indicator function that is equal to 1 if random probability $p>0.5$ (not dropped) in the current training iteration and 0 otherwise.

\paragraph{Representation alignment regularization}  
To accelerate the training process, we adopt a representation alignment regularization term from REPA \cite{yu2024repa}. This term aligns patch-wise projections of the model’s hidden states with a pretrained self-supervised
visual representation, thus providing meaningful guidance to accelerate the model's convergence. Specifically, we use DINOv2 \cite{oquab2023dinov2} as the underlying presentation to provide guidance. Please refer to REPA \cite{yu2024repa} for more details about this term.

The alignment regularization is defined as the maximization of patch-wise similarity between the DINOv2 feature \( f_{d}(\mathcal{I}_r) \) and the projected hidden states \( h_\phi(\mathbf{h}_t) \):

\[
\mathcal{L}_{\text{reg}}(\theta, \phi) := - \mathbb{E}_{\mathbf{x}^0, \epsilon, t} \left[ \frac{1}{N} \sum_{n=1}^{N} \text{sim}\left( f_{d}(\mathcal{I}_r)^{[n]}, h_\phi(\mathbf{h}_t^{[n]}) \right) \right]
\]

where \( n \) indexes each patch, and \( \text{sim}(\cdot, \cdot) \) is a similarity function (e.g., cosine similarity).

\paragraph{Total Loss}  
The total training objective combines the velocity loss and the alignment regularization, which allows robust and efficient training of MMGen:

\[
\mathcal{L}_{\text{total}} = \mathcal{L}_{\text{v}} + \lambda \mathcal{L}_{\text{reg}}
\]

where \( \lambda \) is a weighting factor that balances the contribution of the alignment regularization.

\section{Experiments}

\subsection{Implementation}
We follow the setup described in the SiT and DiT frameworks, using the ImageNet-1k \cite{deng2009imagenet} dataset preprocessed to a resolution of 256×256. Each image is encoded into a compressed latent representation \(x \in \mathbb{R}^{32 \times 32 \times 4} \) using the pretrained Stable Diffusion VAE \cite{rombach2022latentdiff}. For model configurations, we utilize XL/2 architecture, as in the SiT \cite{ma2024sit} and REPA \cite{yu2024repa} setups, with a consistent patch size of 2. The model is trained on 8 NIVDIA A100 GPUs for about two days. The training batch size is set to 256. 
Following SiT \cite{ma2024sit} and REPA\cite{yu2024repa}, we utilize the SDE Euler-Maruyama sampler (for SDE with \( w_t = \sigma_t \)) to generate 50,000 samples and set the default number of function evaluations (NFE) to 250. We report Fréchet Inception Distance (FID \cite{heusel2017gans_fid}) and sFID
\cite{nash2021generating_sfid} for quantitative evaluations. 
Please refer to the supplementary material for more implementation details.

\subsection{Multi-Modal Generation}
To perform multi-modal generation during inference, we apply a single time scheduler across all modalities, using the same timestep $t$ throughout the diffusion process. This allows for the simultaneous generation of multiple modalities conditioned on the specified category. Following this strategy, we randomly generate 50,000 samples and compare MMGen with SiT and REPA under the category-conditioned generation setting.
Table \ref{tab:mmgen_eval_rgb} presents quantitative comparisons of generated RGB using FID and sFID metrics. Our model achieves comparable performance to REPA, reaching similar quality with only a limited number of additional iterations, while also converging significantly faster than SiT. Notably, both SiT and REPA are trained solely on the RGB modality. Please refer to Fig. 3 in the supplementary material for qualitative results.

The training efficiency of MMGen compared to SiT can be attributed to the guidance provided by the representation alignment regularization, which enhances convergence speed. In comparison to REPA, we argue that the additional challenge posed by incorporating multi-modality information in training accounts for the slight increase in iterations, as it introduces greater complexity to the learning process.

\begin{table}[ht!]
\centering\small
\caption{Qualitative comparisons with baseline methods. All results are reported without classifier-free guidance.
}
\begin{tabular}{lcccc}
\toprule
     Model & \#Params & Iter. & FID$\downarrow$ & sFID$\downarrow$ \\
     \midrule
     SiT-XL/2 \cite{ma2024sit}  & 675M & 7M   & 8.3 & 6.32 \\

     REPA \cite{yu2024repa} & 675M & {400K}  & 7.9 & 5.06  \\
     Ours & 695M & {400K} & 9.8 & 5.25 \\
     Ours & 695M & {600K} & \textbf{7.8} & \textbf{4.90}  \\
\bottomrule
\end{tabular}
\label{tab:mmgen_eval_rgb}
\end{table}

\subsection{Multi-modal conditioned Generation}

In this section, we evaluate MMGen’s ability to multi-modal generation from fine-grained conditions, such as depth maps. During inference, we adopt different time schedulers for the condition modality (\( t \in [0.99, 1] \)) and the other modalities (i.e., RGB and others, \( t \in [0, 1] \)). This strategy ensures that the condition modality retains its information when blended with random noise, while the other modalities are generated starting from random noise. Enhanced with the corresponding task embedding \( e_t \) for different condition modalities, our model is capable of performing multi-modal conditioned generation effectively.

\begin{figure}
    \centering
    \begin{overpic}[width=\linewidth]{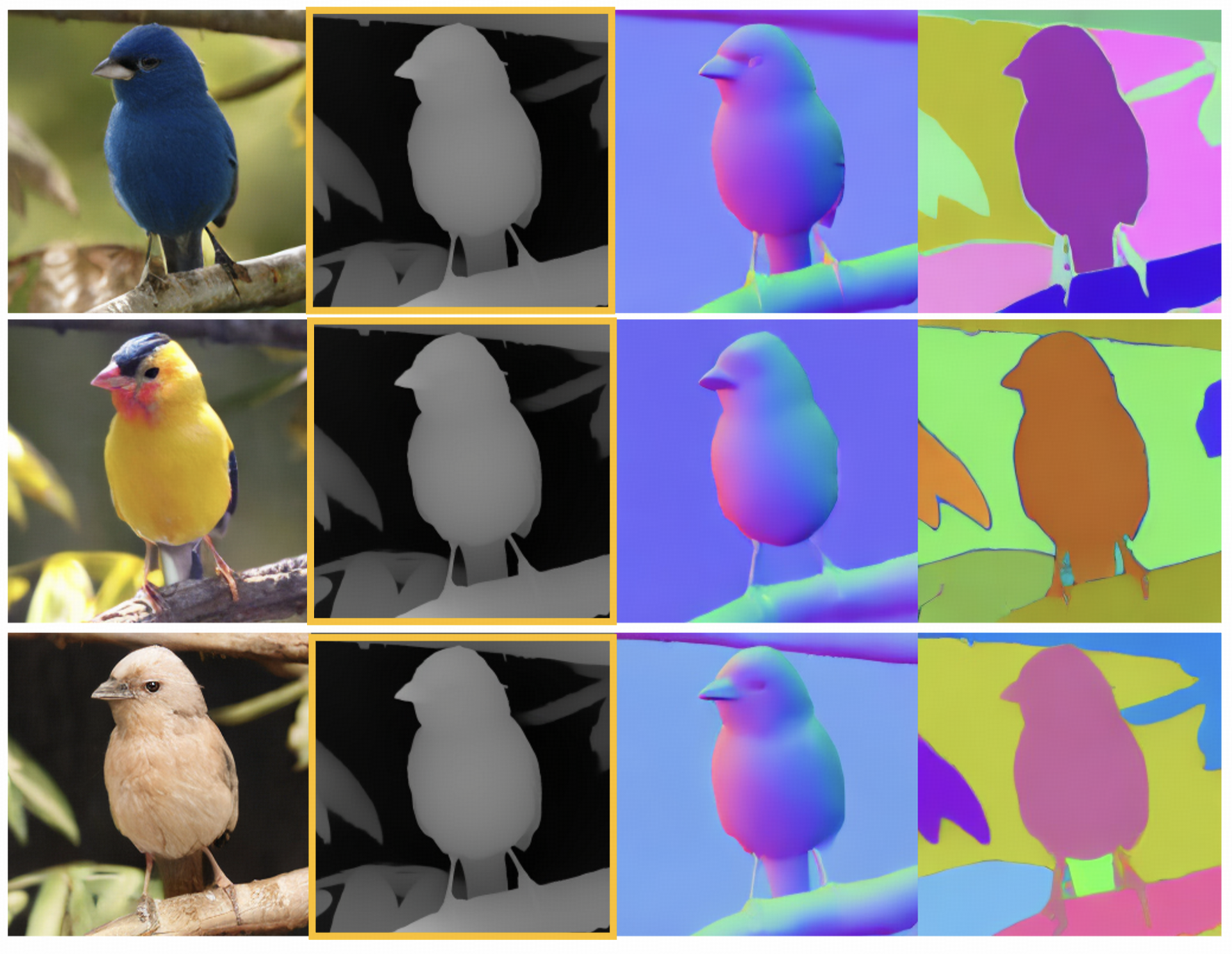}
        \put(5,-2){\small (a) rgb}
        \put(30,-2){\small (b) depth}
        \put(55,-2){\small (c) normal}
        \put(80,-2){\small (d) mask}
    \end{overpic}
    \caption{\textbf{Diversity of depth-conditioned generation.} Given the same depth condition, MMGen can generate diverse RGB images and other aligned modalities.}
    \vspace{-12pt}
    \label{fig:cond_gen_depth_div}
\end{figure}
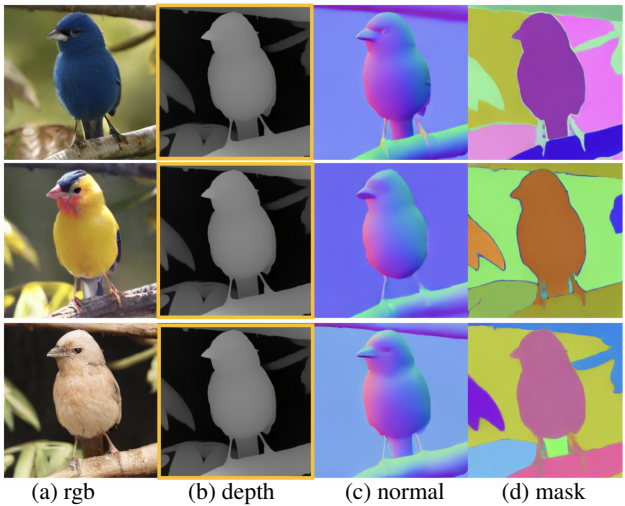

Since this feature is not supported by REPA and SiT, we use the powerful ControlNet \cite{zhang2023controlnet} as a baseline to assess MMGen's performance in conditioned generation.
It is important to highlight that ControlNet is trained and finetuned on extensive, high-quality datasets, whereas MMGen is trained from scratch only on ImageNet-1k. While this difference limits direct comparisons, we include quantitative results to provide insights into MMGen’s capabilities. 
Table \ref{tab:cond_gen_mm_eval} presents the conditional generation performance on the validation set of ImageNet-1k. Our results show that MMGen achieves much better results over ControlNet.  Additionally, Fig. \ref{fig:cond_gen_depth_div} shows that our method can produce diverse images given the same depth condition. Please refer to Fig. 4, 5, and 6 in the supplementary for visualization results.

It is important to note that ControlNet requires fine-tuning separate models for different modalities, whereas our approach utilizes a single diffusion model. Additionally, ControlNet is limited to only generate RGB images based on conditions, while MMGen can generate multiple modalities simultaneously. For instance, given a depth condition, MMGen can produce a corresponding RGB image, along with normal maps and segmentation masks, providing a more comprehensive and versatile output.
\begin{table}[ht!]
\centering\small
\caption{\textbf{Quantitative evaluation of conditioned generation}. ControlNet-D, ControlNet-N, ControlNet-M indicate the ControlNet model is finetuned on depth, normal, and mask conditions, respectively. Our method uses a single unified model.  Note that we use classifier-free guidance with $w=1.8$ for our method.}
\begin{tabular}{cccccccc}
\toprule
     Model & FID$\downarrow$ & sFID$\downarrow$   \\

\midrule
     ControlNet-D \cite{zhang2023controlnet}   & 13.6 & 12.5 \\
     Ours        & \textbf{3.7}   & \textbf{4.2}  \\
\midrule
     ControlNet-N \cite{zhang2023controlnet}   & 19.1 & 15.4  \\
     Ours        &  \textbf{4.6}  & \textbf{4.3} \\
\midrule
     ControlNet-M \cite{zhang2023controlnet}   & 16.1 & 16.6 \\
     Ours        & \textbf{5.6}   & \textbf{4.4}  \\
\bottomrule
\end{tabular}
\label{tab:cond_gen_mm_eval}
\end{table}

\subsection{Multi-Modal Visual Understanding}
Benefiting from our unified framework, MMGen possesses multi-modal understanding capabilities, enabling it to generate multiple visual modalities within a single diffusion process from an input image. However, evaluating our method poses challenges. On the one hand, there are no suitable baselines capable of performing the same task—simultaneously generating multiple visual modalities in a single diffusion process. On the other hand, MMGen is a category-conditioned model trained on ImageNet, making it difficult to directly evaluate on commonly used benchmarks such as ScanNet. Nevertheless, to gain insights into MMGen’s performance, we first conduct quantitative comparisons on the ImageNet validation set. Meanwhile, to further assess MMGen’s quantitative capabilities, we evaluate its depth estimation performance on the indoor scene dataset ScanNet \cite{dai2017scannet}, despite the lack of direct overlap between ImageNet categories and indoor environments.

To evaluate the visual understanding task, we use the RGB modality as condition with a time scheduler (\( t \in [0.99, 1] \)) while applying \( t \in [0, 1] \) to the other modalities. This setup, along with the corresponding task embedding \( e_t \), enables MMGen to perform multi-modal visual understanding tasks effectively. 
Fig. \ref{fig:vis_scannet} and Fig. 7, 8 in the supplementary material present qualitative generation results on ScanNet and ImageNet. These results demonstrate our model's ability to understand visual properties of depth, normal, and segmentation simultaneously while ensuring consistency with the input image observations.

To conduct a quantitative evaluation of MMGen’s zero-shot performance on visual understanding, we randomly selected 5, 000 images from the ScanNet dataset \cite{dai2017scannet} and used each RGB image as conditioning input. MMGen then generated multi-modal understanding results in a single diffusion pass. Here, we adopted one modality, i.e., depth, and compare MMGen’s predictions with those from the widely used diffusion-based method Marigold \cite{ke2023marigold}, specifically designed for depth estimation, alongside the ScanNet ground truth. Table \ref{tab:evaluation_normal} presents quantitative comparisons.
Our method can achieve comparable performance with Marigold, demonstrating the effectiveness of MMGen.
Please refer to the supplementary material for more visualization results.

\begin{table}[ht!]%{0.5\linewidth}
    \centering
    \caption{\textbf{Quantitative evaluation of depth estimation.}}
    \begin{tabular}{ccccccccc}
        \hline
         Method & AbsRel$\downarrow$ & $\delta1$ $\uparrow$  & RMSE $\downarrow$ \\
        \hline
        Marigold \cite{ke2023marigold} & 0.080 & \textbf{0.930}  &  \textbf{0.201} \\
        Ours  & \textbf{0.079} & \textbf{0.930} & 0.226\\
        \hline
    \end{tabular}
    \vspace{-8pt}
    \label{tab:evaluation_normal}
\end{table}

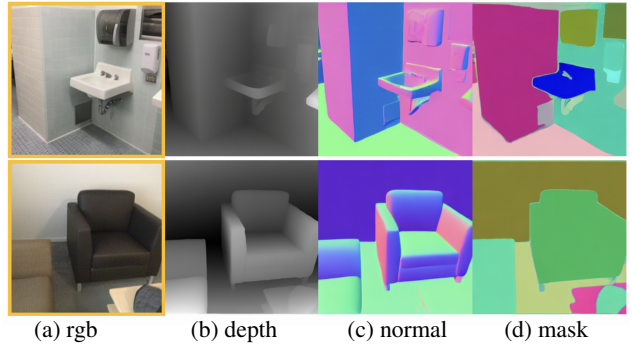
\begin{figure}
    \centering
    \begin{overpic}[width=\linewidth]{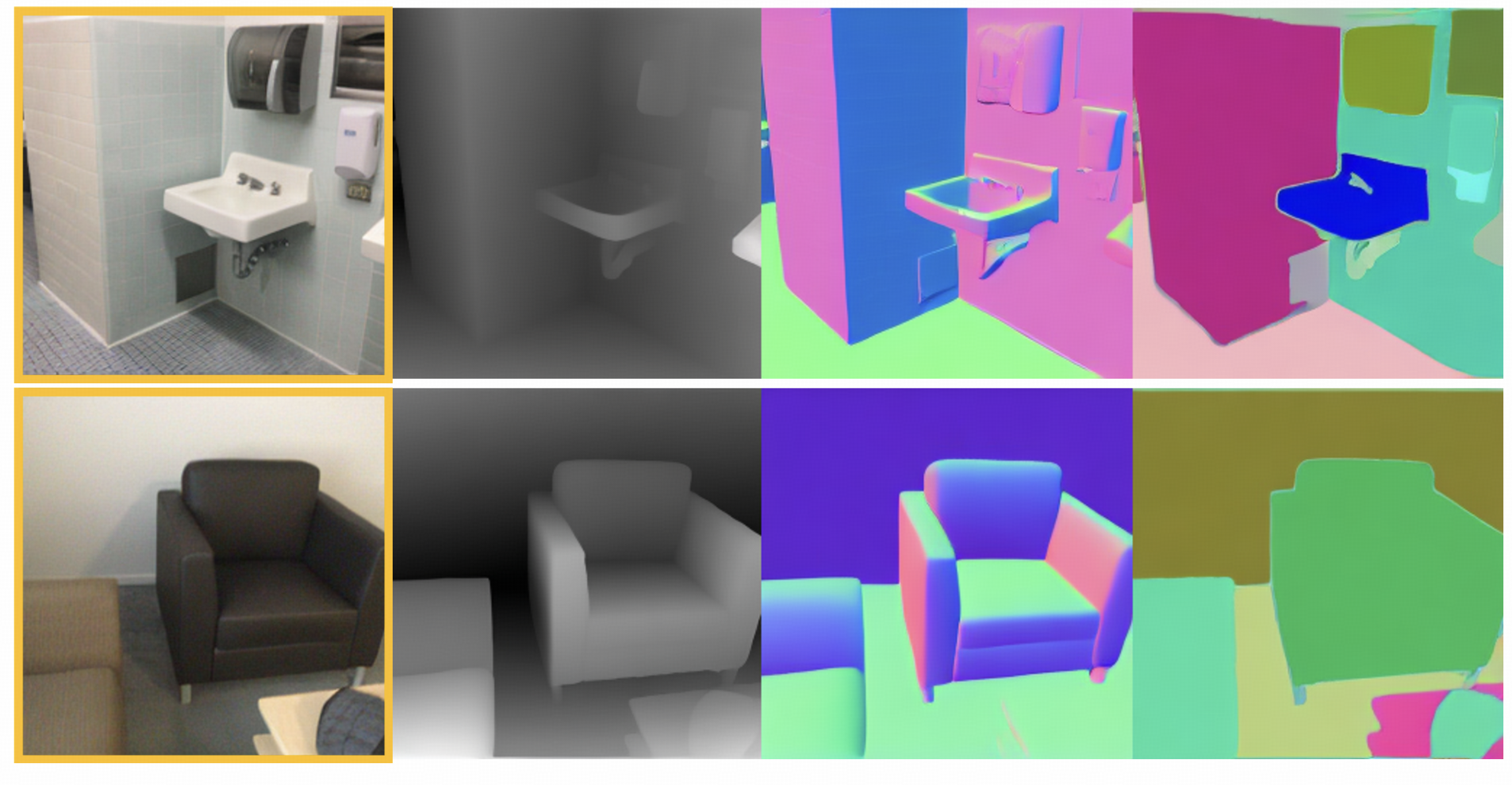}
        \put(5,-2){\small (a) rgb}
        \put(30,-2){\small (b) depth}
        \put(55,-2){\small (c) normal}
        \put(80,-2){\small (d) mask}
    \end{overpic}
    \caption{\textbf{Qualitative results of visual understanding on ScanNet}. Conditioned on rgb (a), MMGen can predict the associate depth, normal and mask simultaneously.}
    \label{fig:vis_scannet}
\end{figure}

\subsection{Ablation}
To ablate the effectiveness of each module, we conduct experiments with four different configurations: (1) Ours-Gen: training only for a multi-modal category-conditioned generation; (2) w/o augmentation: training with an augmentation strategy via batch mixing, where half batch only supervise RGB, and half batch supervision for normal, depth, and mask tasks is randomly omitted. Besides, in the second half, only one quarter is adopted for task decoupling. (3) w/o task embedding: removing the task enhancement signal from the conditioning inputs; and (4) Ours-full: our full setting. Table \ref{tab:ablation} summarizes the quantitative results for each setting. Our full setting can achieve the best FID and sFID, showing the effectiveness of each module. 
\begin{table}[ht!]
\centering\small
\caption{\textbf{Ablation study}. 
Each setting is evaluated after 400K iterations without classifier-free guidance on RGB modalities.
}
\begin{tabular}{ccccccccc}
\toprule
     Model & FID$\downarrow$ & sFID$\downarrow$ & MMGen & CondGen & Vis\\
\midrule
     Ours-Gen & 11.4  & 5.6 & $\checkmark$ & $\times$ & $\times$\\ 
\midrule
     w/o aug & 11.6 & 5.9  & $\checkmark$ & $\checkmark$ & $\checkmark$\\
     w/o T-emb & 12.6 & 5.8 & $\checkmark$ & $\checkmark$ & $\checkmark$\\
     Ours-full & \textbf{9.8} & \textbf{5.3} & $\checkmark$ & $\checkmark$ & $\checkmark$\\
\bottomrule
\end{tabular}
\label{tab:ablation}
\end{table}

\subsection{Applications}\label{sec:application}

\paragraph{Image-to-image translation} MMGen can be utilized for image-to-image translation. Given a reference image, MMGen can interpret it into three visual modalities simultaneously. Then, for each modality, we can feed into MMGen again as conditions to generate a new image. 
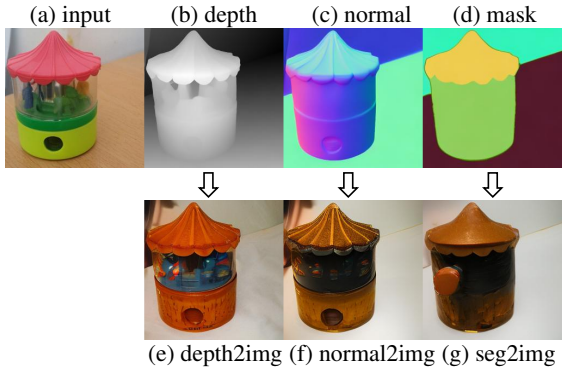
\begin{figure}
    \centering
    \begin{overpic}[width=0.9\linewidth]{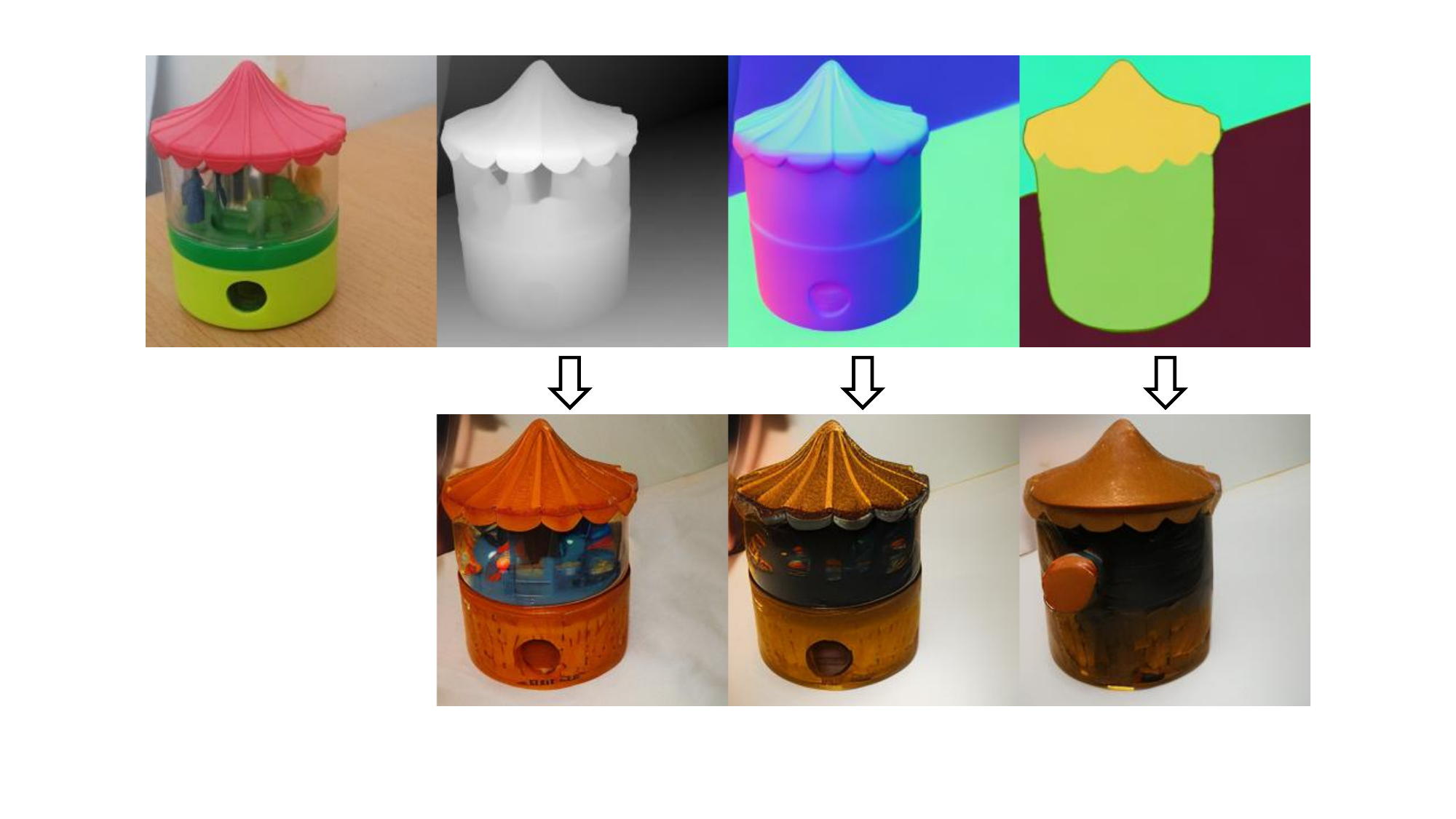}
        \put(5,57){\small (a) input}
        \put(30,57){\small (b) depth}
        \put(55,57){\small (c) normal}
        \put(80,57){\small (d) mask}

        \put(26,-3){\small (e) depth2img}
        \put(51,-3){\small (f) normal2img}
        \put(78,-3){\small (g) seg2img}
    \end{overpic}
    \caption{
    \textbf{Image-to-image translation}. Given input image (a), MMGen predicts (b,c,f) in one diffusion process. Then, for each condition, MMGen can perform conditioned generation to get a novel image respectively (b$\rightarrow$e, c$\rightarrow$f, d$\rightarrow$g).
    }
    \vspace{-12pt}
    \label{fig:img2img}
\end{figure}

\begin{figure}
    \centering
    \begin{overpic}[width=\linewidth]{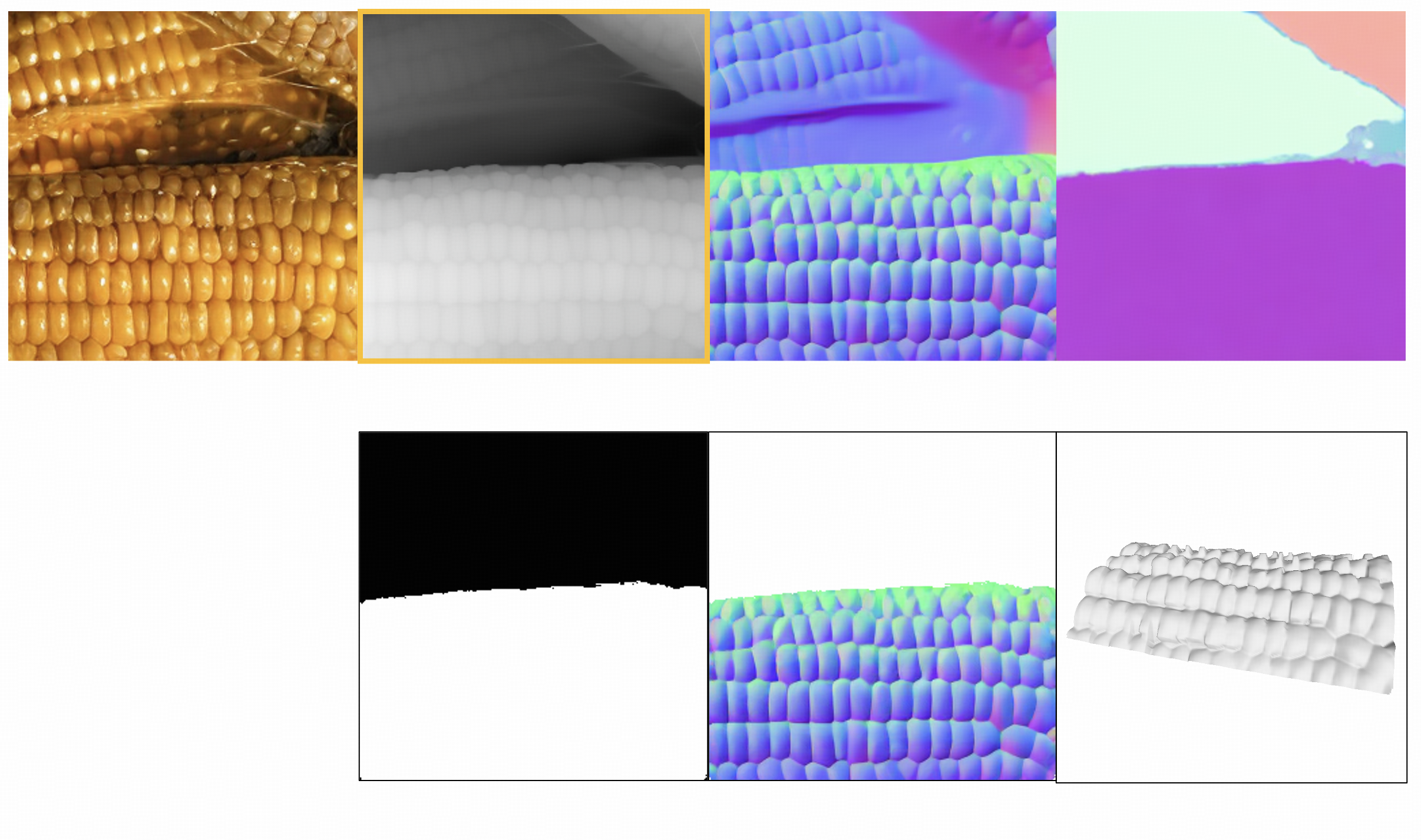}
        \put(5,30){\small (a) rgb}
        \put(30,30){\small (b) depth}
        \put(55,30){\small (c) normal}
        \put(80,30){\small (d) mask}

        \put(29,0){\small (e) selection}
        \put(53,0){\small (f) masked n-}
        \put(80,0){\small (g) mesh}
            
    \end{overpic}
    \caption{\textbf{3D reconstruction via MMGen}. Starting from a depth map (b), our method generates a high-quality RGB image (a), an aligned normal map (c), and semantic segmentation results (d). The purple region in (d) is used as a mask (e) to extract the masked normal map (f). Then, (e) and (f) serve as inputs for downstream mesh reconstruction (g). }
    \label{fig:3drecon}
\end{figure}
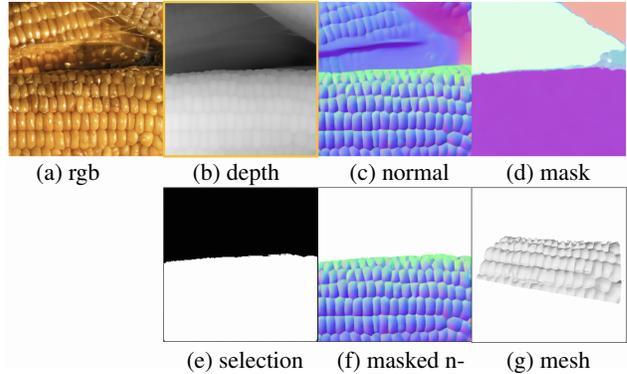

\paragraph{3D Reconstruction} 
MMGen can be used for 3D reconstruction of foreground objects without the need to run an individual segmentation model.
As shown in Fig. \ref{fig:3drecon}, given a depth map (b), our method can generate other modalities simultaneously (a,c,d). We then select the purple region in (d) as a mask to extract the foreground object (e,f), which serve as inputs for downstream mesh reconstruction via BNI \cite{bini2022normalrecon} (Fig. \ref{fig:3drecon}(g)).
While this task can be performed with separate models—such as using ControlNet for depth-conditioned generation, StableNormal for normal maps, and SemanticSAM for segmentation masks—this approach incurs significant memory and computational costs, as each of these large foundation models operates independently, resulting in approximately three times the cost of MMGen. In contrast, MMGen unifies these capabilities \emph{in only one diffusion process}, reducing memory usage and inference time. 
\vspace{-8pt}

\begin{figure}
    \centering
      \begin{overpic}[width=0.9\linewidth]{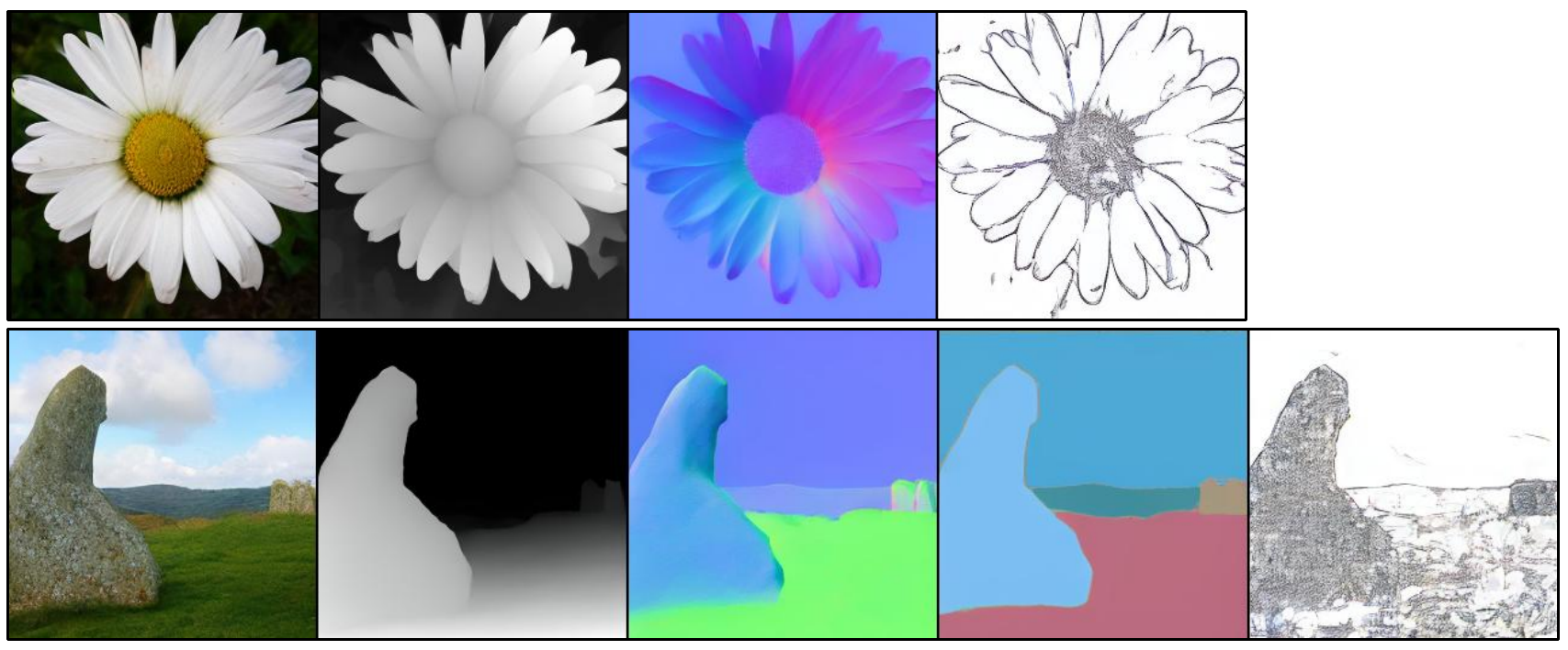}
          \put(-4, 23){\rotatebox{90}{\small (a)ft2canny}}
          \put(-4, 0){\rotatebox{90}{\small (b)add canny}}

        \put(6,42){\small rgb}
        \put(26,42){\small depth}
        \put(45,42){\small normal}
        \put(65,42){\small canny}

        \put(6,-3){\small rgb}
        \put(26,-3){\small depth}
        \put(45,-3){\small normal}
        \put(65,-3){\small mask}
        \put(85,-3){\small canny}
        
      \end{overpic}
    \caption{
    \textbf{Adaptation to new modalities}. (a) Finetune an existing modality to a new modality (seg $\rightarrow$canny); (b) Add an additional modality to support generation of 5 modalities simultaneously.
    }
    \vspace{-16pt}
    \label{fig:ft2canny}
\end{figure}
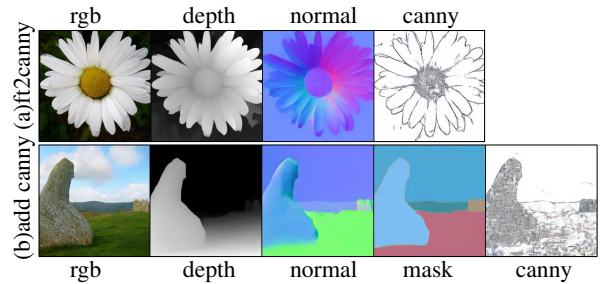

\paragraph{Adaptation to New Modality}
To assess the feasibility of extending MMGen to new modalities, we conducted two experiments using a commonly used modality—Canny edge: 
(1) fine-tuning one existing modality (i.e., segmentation) to Canny with only 1k steps (Fig. \ref{fig:ft2canny} (a)), 
and (2) adding an additional modality to MMGen and fine-tuning it for 10k steps (Fig. \ref{fig:ft2canny} (b)).
These examples demonstrate that our model can be easily adapted to new modalities.

These examples highlight the flexibility of MMGen for various downstream applications within a unified model, as well as its effectiveness in achieving multi-modal generation within a single diffusion process. Furthermore, the Canny fine-tuning experiments show that our model can be easily adapted to new modalities, showcaing potential for seamless integration and expansion across diverse tasks.

\section{Conclusion}

In this paper, we present MMGen, a unified framework for multi-modal generation and understanding that supports multiple tasks within a single model, including multi-modal category-conditioned, fine-grained conditioned generation, and visual understanding. MMGen introduces a multi-modal diffusion transformer building on SiT and a modality-decoupling strategy to achieve synchronized and decoupled multi-modal outputs, demonstrating competitive performance against established models while supporting diverse, simultaneous modalities. Although the model relies on pseudo labels and has limited training resources, future expansions in dataset size and fine-tuning for specific domains hold promise for enhanced performance. As a first step toward a unified multi-modal framework for diffusion-based generation and understanding, we hope MMGen can inspire the development of scalable, versatile AI systems capable of integrated, cross-modal synthesis.

{
    \small
    \bibliographystyle{setup/ieeenat_fullname}
    \bibliography{main}
}

\clearpage
\setcounter{page}{1}
\setcounter{section}{0}
\setcounter{figure}{0}
\maketitlesupplementary

\section{Implementation Details}
\subsection{Data preparation}\label{sec:data}
To train MMGen for multi-modal generation and understanding, we require an aligned multi-modal dataset, denoted as \( \mathcal{I}_{\text{mm}} \), which includes RGB images, depth maps, normal maps, and semantic segmentation masks. Since fully aligned multi-modal datasets are scarce, we generate \( \mathcal{I}_{\text{mm}} \) by creating pseudo-labels from the ImageNet-1k dataset \cite{deng2009imagenet}, utilizing pre-trained 2D foundation models pretrained on large-scale datasets.

Formally, the aligned multi-modal dataset can be represented as:
\[
\mathcal{I}_{\text{mm}} = \{ (\mathcal{I}_r, \mathcal{I}_d, \mathcal{I}_n, \mathcal{I}_s) \mid \mathcal{I}_r \in \text{ImageNet-1k} \},
\]
where \( \mathcal{I}_r \) is the RGB image, \( \mathcal{I}_d \) the depth map from DepthAnythingV2 \cite{depth_anything_v2}, \( \mathcal{I}_n \) the normal map from StableNormal \cite{ye2024stablenormal}, and \( \mathcal{I}_s \) the segmentation mask from SemanticSAM \cite{li2023semanticsam}.

Since our method requires multi-modal inputs, encoding these modalities online during optimization introduces significant computational overhead, reducing training efficiency. To address this, we pre-process the raw pixels of all modalities into compressed latent vectors using a pretrained VAE encoder \cite{rombach2022latentdiff}, following the approaches in REPA \cite{yu2024repa} and EDM2 \cite{karras2024edm2}. Therefore, we don't use data augmentation during training, which has been shown to have minimal impact on performance in REPA and EDM2. Additionally, we precompute the DINOv2-base features \cite{oquab2023dinov2} of RGB images to further reduce the optimization burden and accelerate the training process.

\subsection{Baselines}  
Our model uniquely achieves a unified framework capable of handling visual understanding, category-conditioned, and conditioned image generation within a single model—an ability that no existing diffusion-based work currently matches. Nevertheless, we compare our model’s performance with the most relevant works for each task. 
For category-conditioned image generation, we compare MMGen against the state-of-the-art SiT \cite{ma2024sit} and REPA \cite{yu2024repa}. For fine-grained conditioned generation (i.e., using depth, normal, or mask as conditions), we evaluate our model’s generation quality on the ImageNet-1k validation set \cite{deng2009imagenet} against ControlNet \cite{zhang2023controlnet}, which supports various conditioning inputs. 
For visual understanding tasks, we compare our method with the widely used method Marigold \cite{ke2023marigold}, providing a comprehensive quantitative assessment.

\subsection{Network design motivation}
\begin{figure}
    \centering
    \vspace{12pt}
    \begin{overpic}[width=\linewidth]{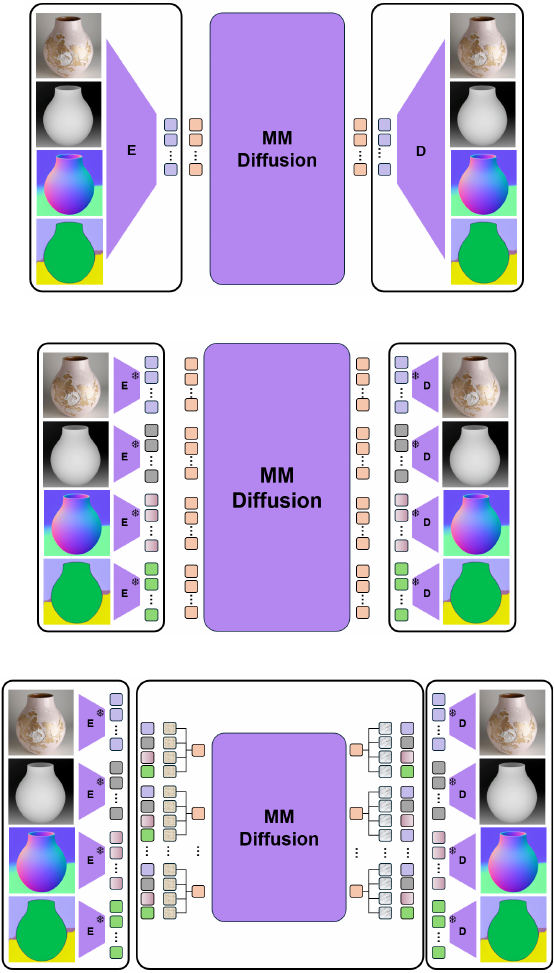}
        \put(0,102){\small \textbf{ (a) Option 1: Joint VAE}. Token number: $N$}
        \put(0,66){\small\textbf{ (b) Option 2: Sequence of image views}. Token number: $4*N$}
        \put(0,31){\small \textbf{ c) Ours: Token fusion}. Token number: $N$}
        
    \end{overpic}
    \caption{
    \textbf{Optional network architecture design}. Note that the orange boxes on the left side of the MM Diffusion block represent the input tokens for transformer diffusion.
    }
    \label{fig:network_design}
\end{figure}

To achieve multi-modal diffusion in one diffusion process, we consider three possible network designs, as illustrated in Fig. \ref{fig:network_design}.  

One option is to train a joint VAE where multiple modalities are encoded into a shared latent space (Fig. \ref{fig:network_design}(a)). This approach maintains a comparable computational cost to standard image diffusion since it does not increase the number of latent tokens (denoted as \( N_{\text{token}} \)). However, once the VAE is trained, these modalities become tightly coupled, making it difficult to decouple them during diffusion and preventing independent control over individual modalities for conditional generation.  

Another option (Fig. \ref{fig:network_design} (b)) is to treat different modalities as a sequence of image views. Each image is encoded using a pretrained VAE, generating latent tokens that are concatenated into a long sequence. For example, with four modalities, the sequence length increases to \( 4 \times N_{\text{token}} \). Given the \( O(n^2) \) complexity of attention mechanisms in transformers, this design significantly increases computational costs—resulting in a 16× higher cost for four modalities—making it impractical under computational constraints.  

To avoid these challenges, we adopt an efficient and flexible modality fusion strategy (Fig. \ref{fig:network_design} (c)). Specifically, we use pretrained encoders to process each modality separately but fuse the modalities at the transformer diffusion input. The outputs are then decoded using separate heads for different modalities. This design allows for flexible integration of additional modalities without requiring a jointly trained VAE. More importantly, after fusion, the token sequence length is reduced to \( N_{\text{token}} \), maintaining a computational cost comparable to standard image diffusion. 

Given these considerations, we opted for this efficient and scalable design.

\subsection{Representation alignment regualarization}
In this section, we discuss the representation alignment regularization (RAR) term used in our framework. In REPA \cite{yu2024repa}, this term aligns self-supervised representations with the diffusion model, where both models are trained exclusively on the RGB modality. In contrast, our diffusion model is trained on multiple modalities.

Theoretically, the optimal approach would involve using a self-supervised model, similar to DINO \cite{oquab2023dinov2}, pretrained on multi-modal images that match the modalities in our diffusion model. However, we were unable to identify a suitable self-supervised model pretrained on all the modalities we use. Despite this limitation, we empirically find that using DINO features for regularization—though suboptimal—significantly accelerates the training of our multi-modal diffusion model. This suggests that there may be inherent connections between multi-modal and RGB representations. Investigating these relationships remains an avenue for future work.

\subsection{Modality decoupling}
As mentioned in the introduction of the main paper, Orchid \cite{wang2024diffx} and \cite{krishnan2025orchid} train joint VAE to encode multiple modalities into a shared latent space, resulting in tightly coupled modalities. This design prevents the use of one modality as a condition to generate others. In contrast, our approach aims to decouple this relationship, enabling greater flexibility. Our method supports various tasks by allowing multiple modalities to be generated simultaneously while also making it possible to use any one of them as a condition to generate the others.

To address the dependencies between different modalities, we propose a modality decoupling strategy, as described in the main paper. The motivation is to enable unified generation and understanding by aligning and decoupling modality relationships (or noise levels) during denoising process. 
As shown in Figure 1 of the main paper, 
in multi-modal generation (Fig. 1a), all modalities are aligned and generated simultaneously from pure noise. However, in visual understanding (Fig. 1b) and conditioned generation (Fig. 1c), a conditioning modality is provided, while other modalities are generated from pure noise, requiring their relationships to be decoupled rather than aligned. This decoupling is essential to support all these tasks within a unified framework.

\begin{figure}
    \centering
    \begin{overpic}[width=\linewidth]{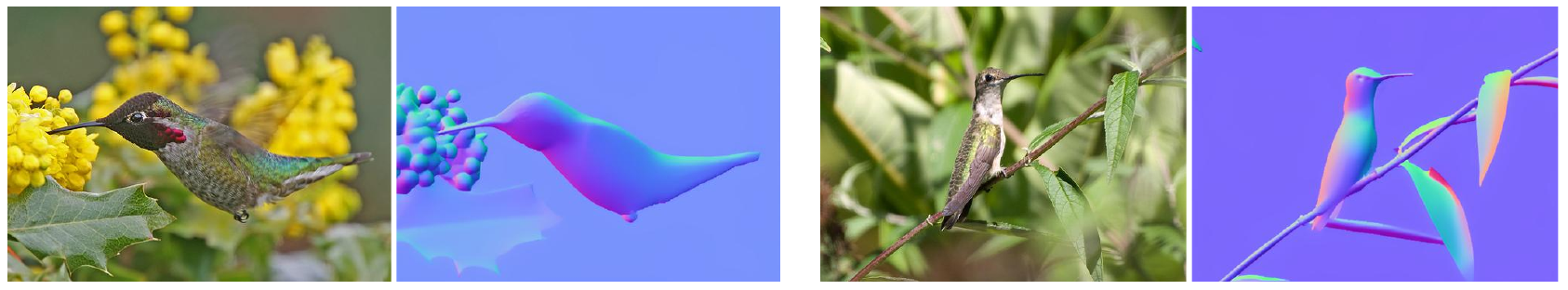}
        
    \end{overpic}
    \caption{ \textbf{Visualization of errors in normal pseudo labels by StableNormal \cite{ye2024stablenormal}.}
    StableNormal struggles to produce accurate estimations in background regions and exhibits variations in reflective areas, such as bird eyes.
    }
    \label{fig:stablenormal}
\end{figure}

During training, we observed that using entirely independent time schedules for each modality results in slower convergence. Although ComboStoc \cite{xu2024combostoc} shows that fully asynchronous time steps for image patches and feature vectors can alleviate insufficient sampling and accelerate training, directly applying this approach to our task for modality decoupling is challenging.
First, ComboStoc \cite{xu2024combostoc} focuses on processing image patches with localized, pixel-level features to address insufficient sampling, while our approach aims to decouple relationships across modalities, which involve more abstract and high-level interactions. The relationships between modalities, including depth, normal, and segmentation, are inherently more complex and distinct compared to the spatial relationships between patches within a single image. Modalities represent different aspects of the scene's semantics and geometry, making their inter-dependencies far more challenging for the model to learn, especially with overly flexible time schedulers.
Moreover, we rely on expert models to generate pseudo labels for training, which contain inherent errors and inconsistencies, as shown in Fig. \ref{fig:stablenormal}. The lack of constraints between modalities could lead the model to overfit these inaccuracies.

To mitigate these challenges, we adopt a more constrained approach by applying a unique time scheduler only for the conditioned modality, while using a shared scheduler for the remaining three modalities. This strategy simplifies the training process, prevents overfitting to noisy pseudo labels, and helps the model converge more efficiently.

\subsection{Shared encoder for all modalities} 
In our design, we adopt a shared image encoder for all modalities.
There are two main considerations. First, 
since all images exist in pixel space, 
using a shared encoder helps distinguish different patches more effectively. 
For separate encoders trained for each modality, 
there is a risk that image patches in different modalities may be mapped to similar codes, making training more challenging. 
Second, a shared encoder significantly reduces training computational costs, as training separate encoders would require substantially more resources. 
While training separate VAEs is a viable alternative, we leave this as a future research direction.

\section{More Results}
This section presents additional results for multi-modal generation and visual understanding.

\subsection{Multi-modal category-conditioned generation}
\begin{figure*}
    \centering
    \begin{overpic}[width=\linewidth]{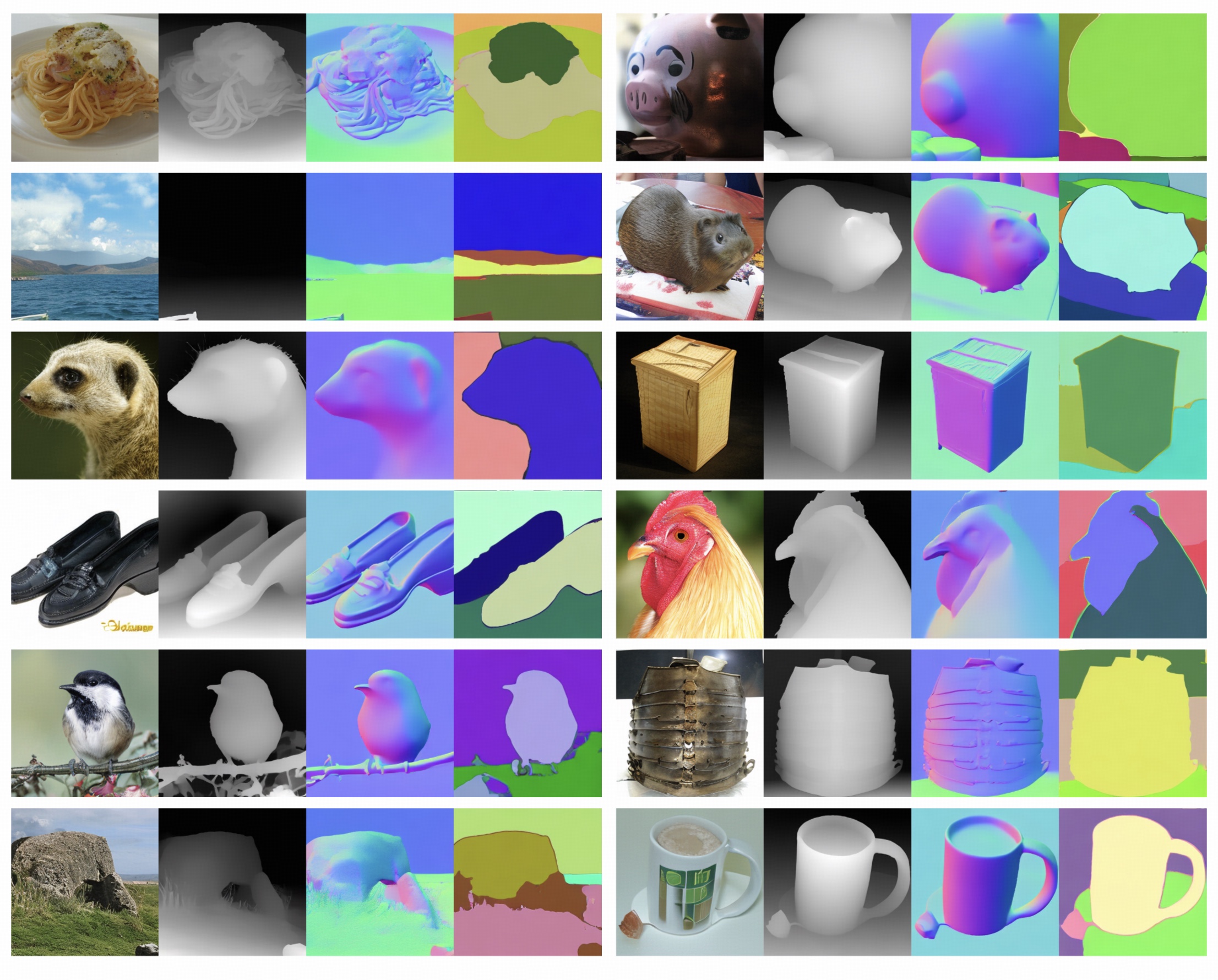}
        \put(3.5,-1){ \small (a) rgb}
        \put(15.5,-1){ \small (b) depth}
        \put(27,-1){ \small (c) normal}
        \put(40,-1){ \small (d) mask}

        \put(53.5,-1){ \small (a) rgb}
        \put(65.5,-1){ \small (b) depth}
        \put(77,-1){ \small (c) normal}
        \put(90,-1){ \small (d) mask}
    \end{overpic}
    \caption{\textbf{Multi-modal category-conditioned generataion}. }
    \label{fig_supp:mmgen}
\end{figure*}

Fig. \ref{fig_supp:mmgen} presents visual examples of multi-modal category-conditioned generation. Our model produces high-quality, diverse, and well-aligned multi-modal outputs within a single diffusion process. This approach eliminates the need for separate models during inference, significantly improving efficiency and reducing computational cost.

\subsection{Multi-modal conditioned generation} 
In the main paper, we compared three conditioned generation results with individual ControlNet \cite{zhang2023controlnet} for each condition. Here, we provide more discussions about the generation resutls.

Table 2 in the main paper presents the quantitative comparisons across the three conditions. 
Compared to the individual models of ControlNet, our unified model achieves superior FID scores across all conditions. Additionally, our method simultaneously generates other aligned outputs, further demonstrating its versatility.
For different conditions in our method, the best FID is achieved on the depth-conditioned setting. We attribute this to the richer information provided by depth conditions, which leads to superior generation performance compared to the other conditions. 
Figs. \ref{fig_supp:condgen_depth}, \ref{fig_supp:condgen_normal}, and \ref{fig_supp:condgen_seg} present qualitative comparisons between our method and ControlNet for depth, normal, and segmentation conditions, respectively.

It's important to note that direct comparisons between these two models involve some inherent limitations.  (1) Training differences: ControlNet leverages a large diffusion model pretrained on a massive and diverse dataset (600M image-text pairs) and fine-tunes it on extensive condition-image datasets. For example, ControlNet-Depth fine-tunes the stable-diffusion-v1.5 model on 3M depth-image-caption pairs. In contrast, our method is trained from scratch on a smaller dataset of 1.2M image pairs. (2) Inference differences: While ControlNet and its pretrained model have been trained on large collections of image pairs, they are not specifically optimized for ImageNet-1k dataset, which may introduce a domain gap and degrade performance when computing FID scores. As shown in Fig. \ref{fig_supp:condgen_depth} (f), the outputs from ControlNet exhibit a style that differs from the reference images (a).
Given these objective circumstances, the results presented in Table \ref{tab:cond_gen_mm_eval} may not fully capture the relative strengths of the two models. Nevertheless, we hope these comparisons provide valuable insights into our model’s performance in fine-grained conditioned generation. We believe that training our model on a larger and more diverse dataset would further enhance its performance.

As noted by the authors of ControlNet, \emph{"Learning conditional controls for large text-to-image diffusion models in an end-to-end way is challenging."} In this work, we take the first step toward addressing this challenge by unifying multiple conditional controls and category-conditioned controls within a single model. We hope this initial effort inspires future research in advancing unified frameworks for multi-modal conditional generation.

\subsection{Multi-modal visual understanding} To verify the effectiveness of our method on visual understanding, we test the visual understanding performance on ImageNet-1k validation set and the generalization ability on the widely used ScanNet \cite{dai2017scannet} dataset. Fig. \ref{fig_supp:vision_imagenet} and Fig. \ref{fig_supp:vision_scannet} show the visual understanding results on ImageNet-1k and ScanNet datasets, respectively.

\section{Limitations and Future Work}
While our model demonstrates strong performance in multi-modal generation and understanding, it has certain limitations. 

\begin{figure*}
    \centering
    \begin{overpic}[width=0.8\linewidth]{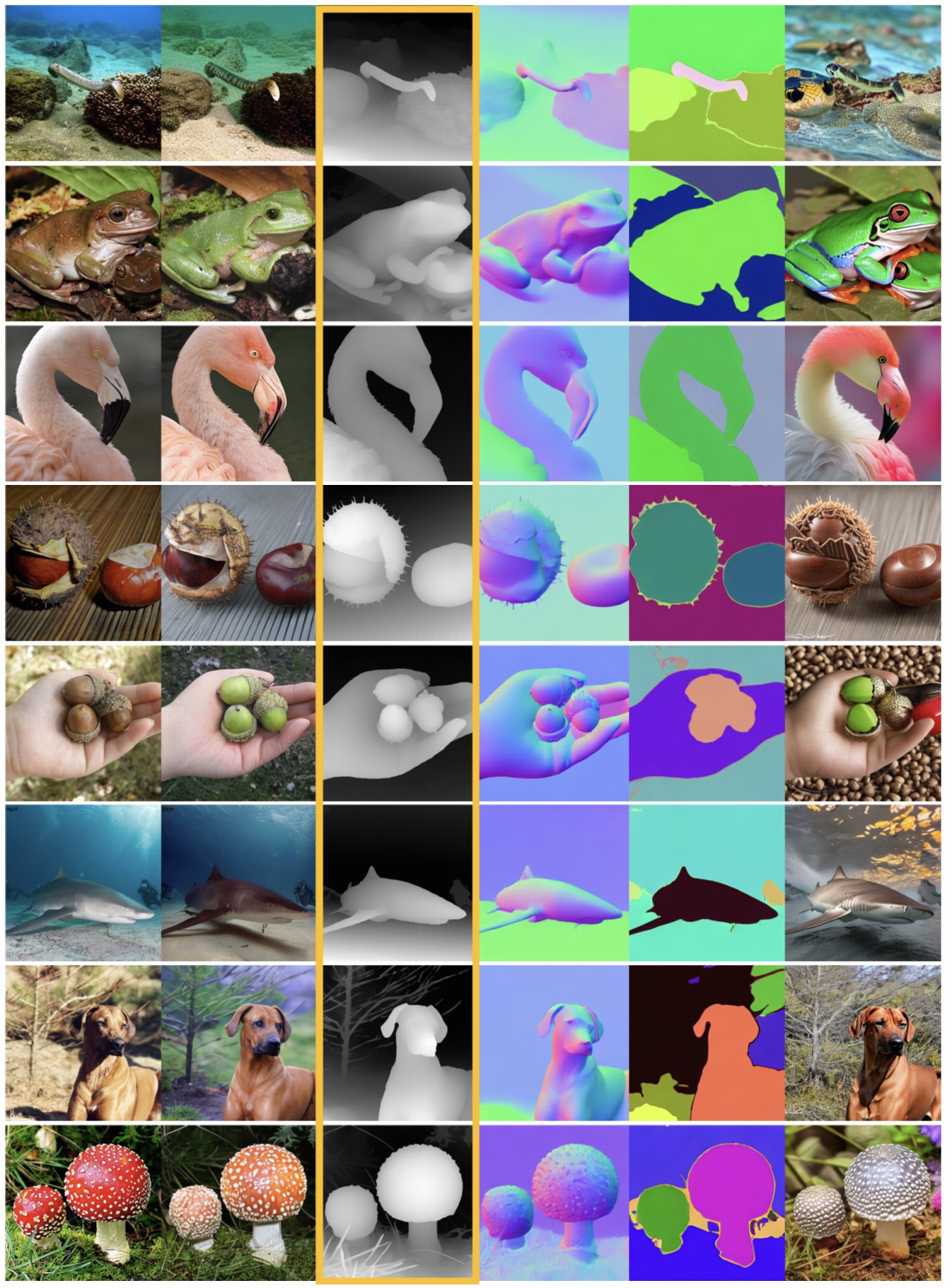}
        \put(2,-1){ \small (a) reference}
        \put(14,-1){ \small (b) rgb}
        \put(27,-1){ \small (c) depth}
        \put(39,-1){ \small (d) normal}
        \put(51,-1){ \small (e) mask}
        \put(62,-1){ \small (f) ControlNet}
    \end{overpic}
    \caption{Multi-modal \textbf{depth-conditioned} generation. }
    \label{fig_supp:condgen_depth}
\end{figure*}

\begin{figure*}
    \centering
    \begin{overpic}[width=0.8\linewidth]{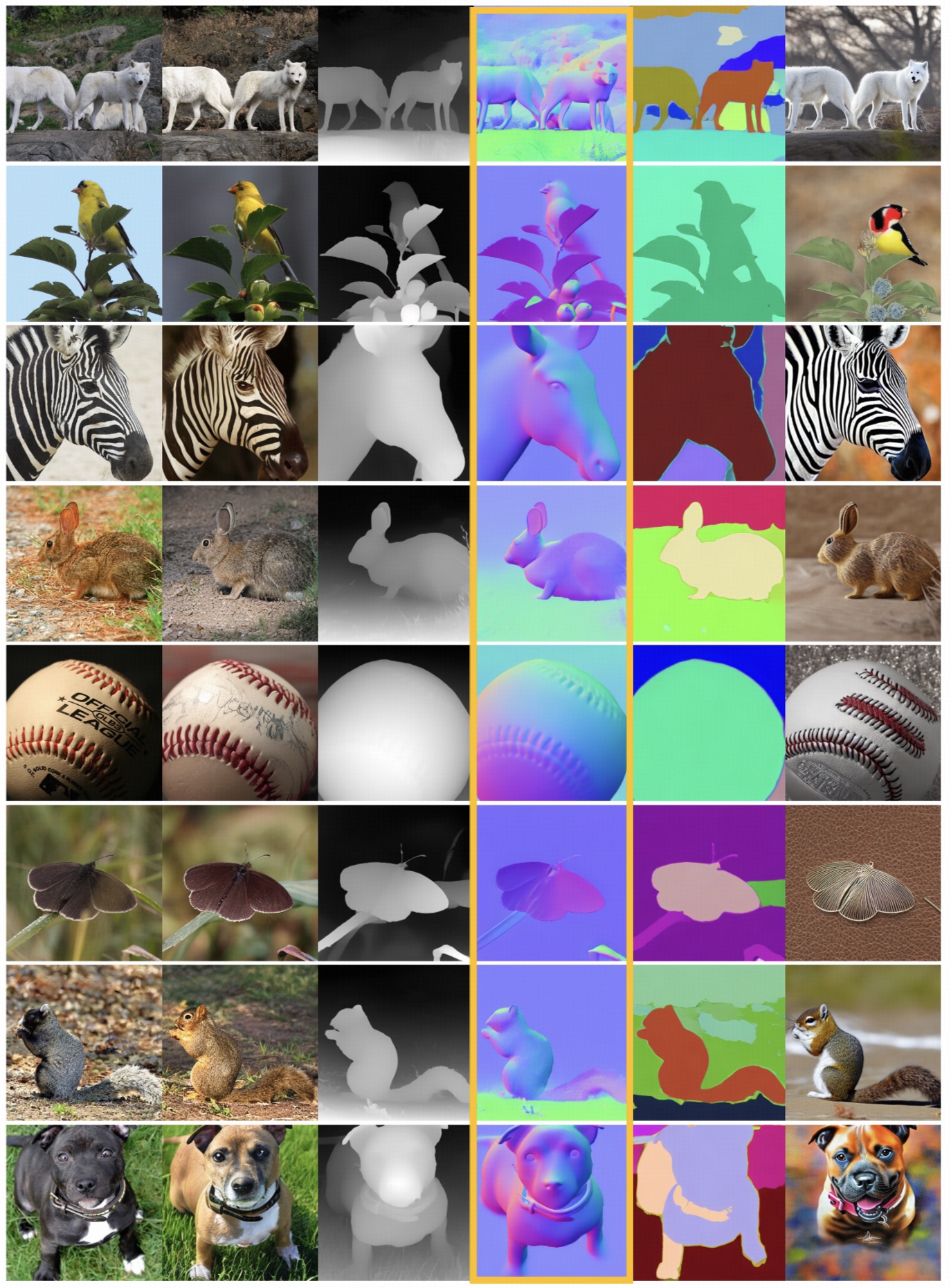}
        \put(2,-1){ \small (a) reference}
        \put(15.5,-1){ \small (b) rgb}
        \put(27,-1){ \small (c) depth}
        \put(39,-1){ \small (d) normal}
        \put(52.5,-1){ \small (e) mask}
        \put(62,-1){ \small (f) ControlNet}
    \end{overpic}
    \caption{Multi-modal \textbf{normal-conditioned} generation. }
    \label{fig_supp:condgen_normal}
\end{figure*}

\begin{figure*}
    \centering
    \begin{overpic}[width=0.8\linewidth]{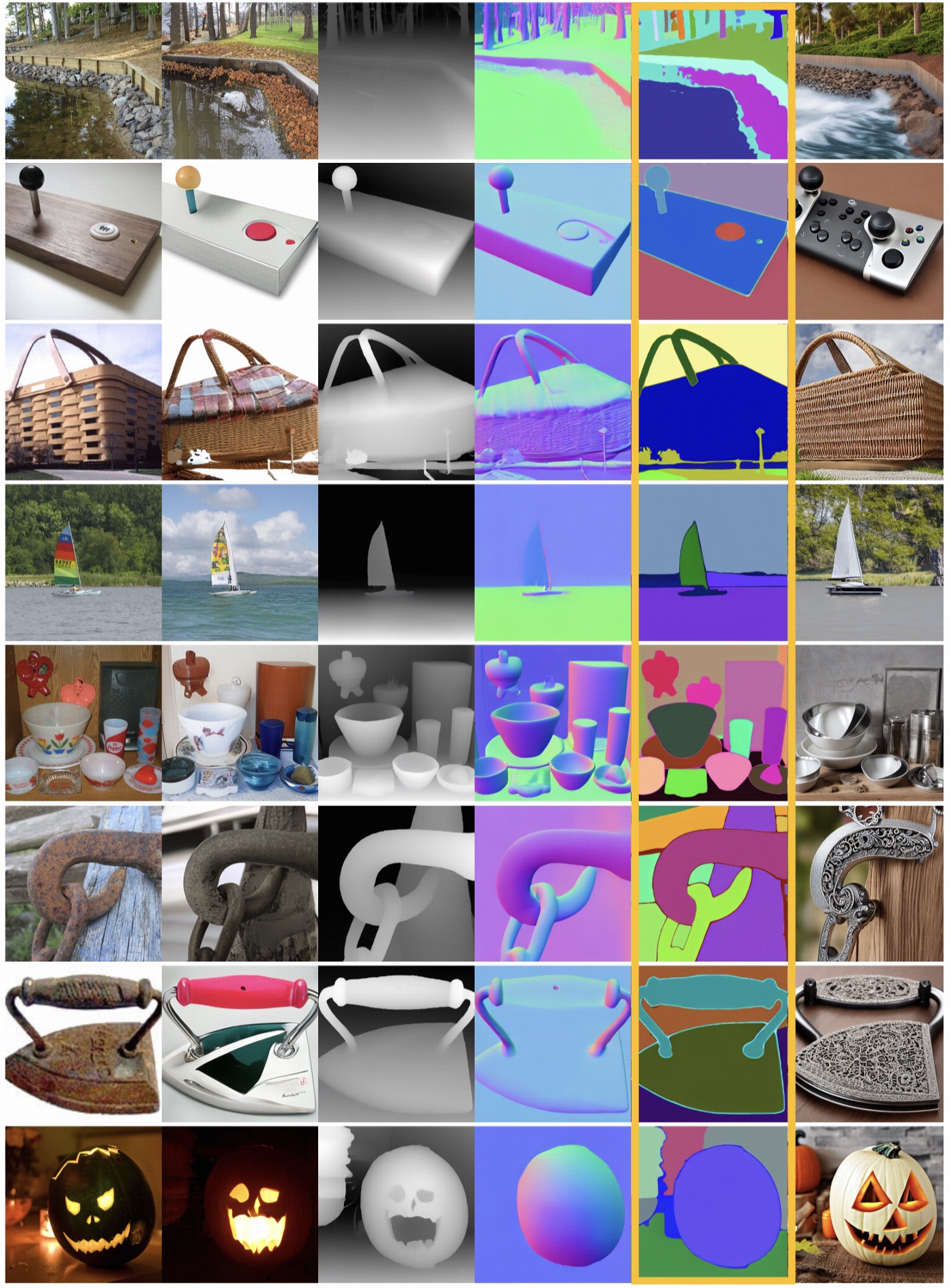}
        \put(2,-1){ \small (a) reference}
        \put(15.5,-1){ \small (b) rgb}
        \put(27,-1){ \small (c) depth}
        \put(39,-1){ \small (d) normal}
        \put(52.5,-1){ \small (e) mask}
        \put(62,-1){ \small (f) ControlNet}
    \end{overpic}
    \caption{Multi-modal \textbf{segmentation-conditioned} generation. }
    \label{fig_supp:condgen_seg}
\end{figure*}

\begin{figure*}
    \centering
    \begin{overpic}[width=\linewidth]{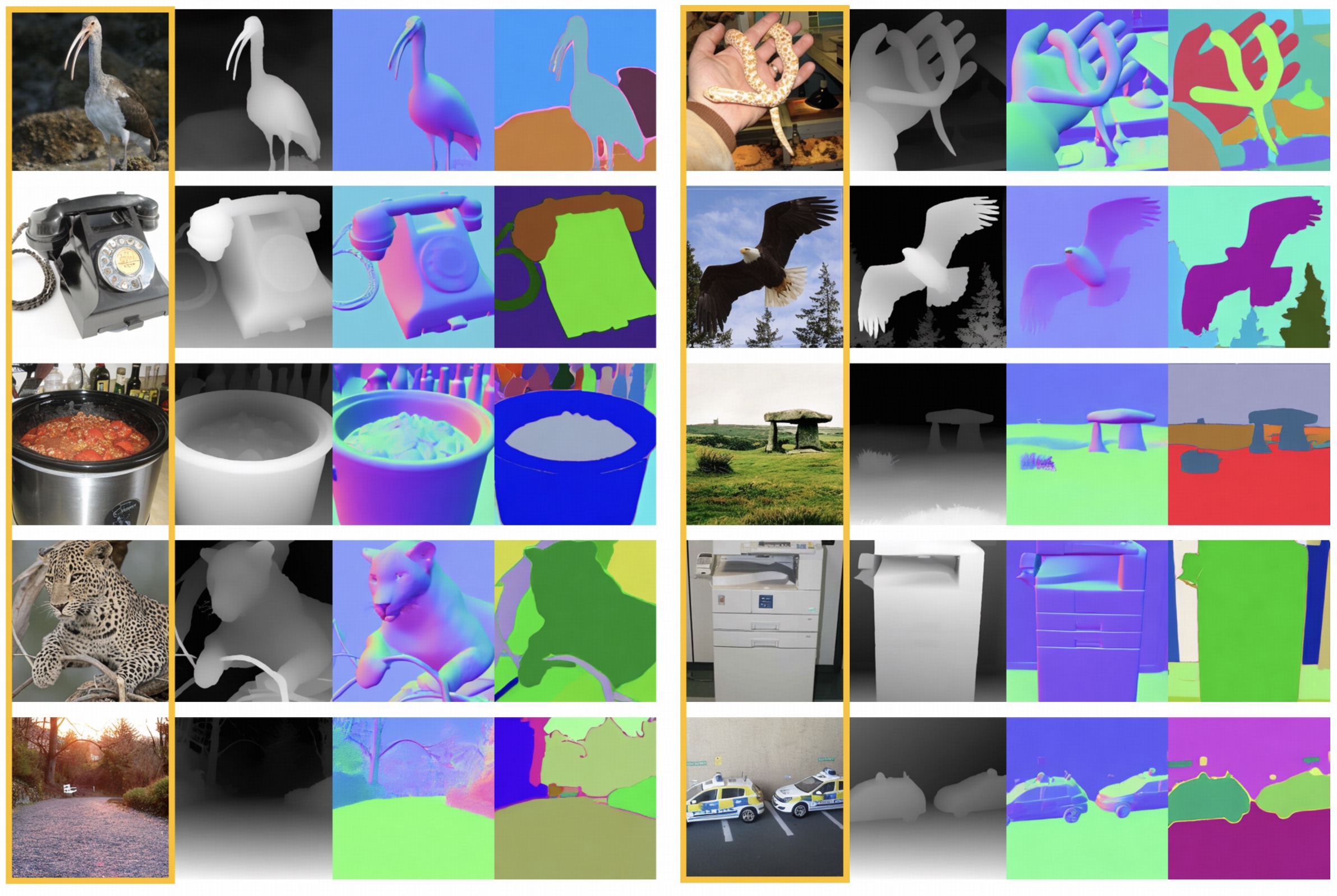}
        \put(3.5,-1){ \small (a) rgb}
        \put(15.5,-1){ \small (b) depth}
        \put(27,-1){ \small (c) normal}
        \put(40,-1){ \small (d) mask}

        \put(53.5,-1){ \small (a) rgb}
        \put(65.5,-1){ \small (b) depth}
        \put(77,-1){ \small (c) normal}
        \put(90,-1){ \small (d) mask}
    \end{overpic}
    \caption{Multi-modal visual understanding on \textbf{ImageNet-1k validation set}. }
    \label{fig_supp:vision_imagenet}
\end{figure*}

\begin{figure*}
    \centering
    \begin{overpic}[width=\linewidth]{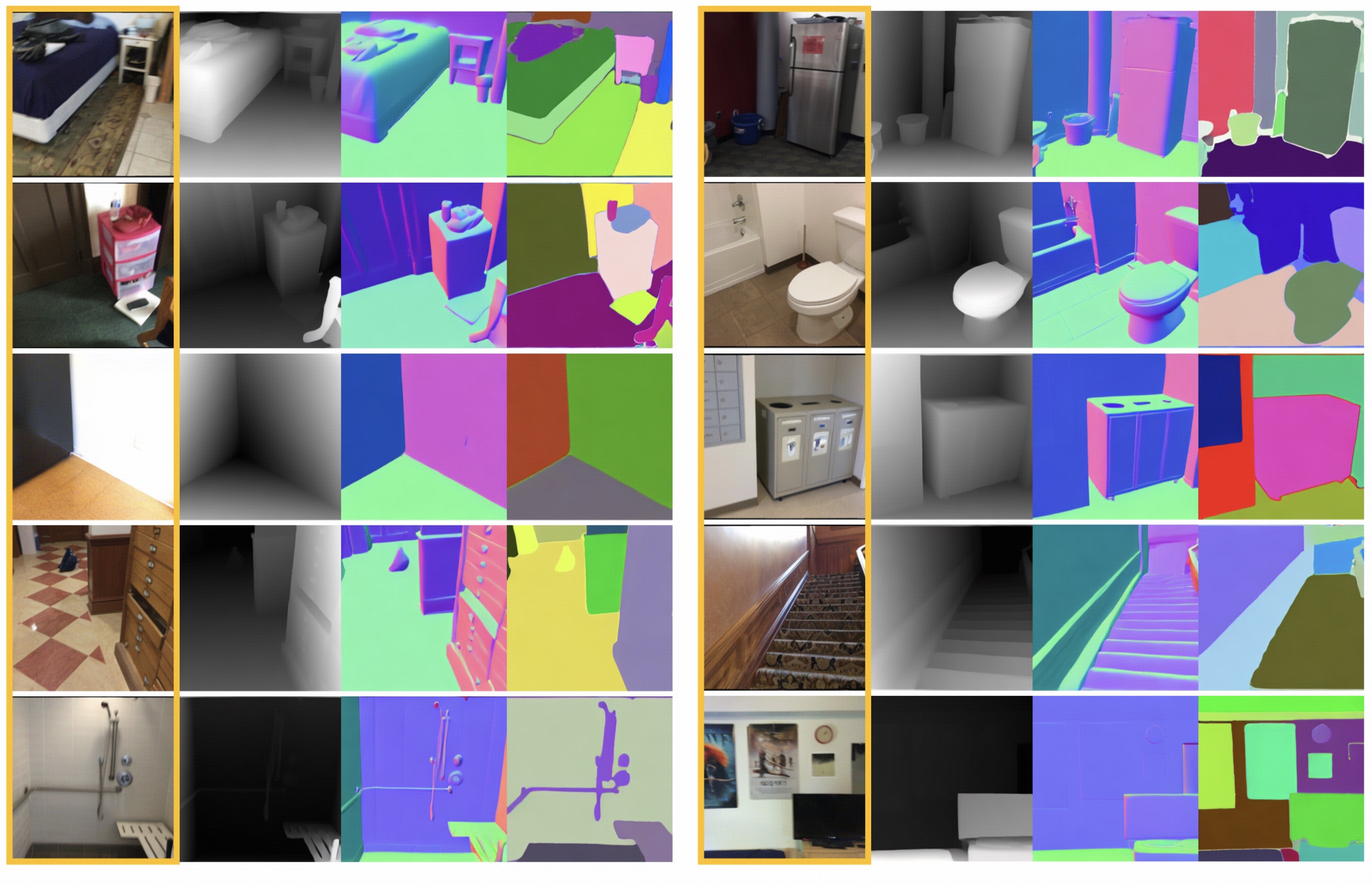}
        \put(3.5,-1){ \small (a) rgb}
        \put(15.5,-1){ \small (b) depth}
        \put(27,-1){ \small (c) normal}
        \put(40,-1){ \small (d) mask}

        \put(53.5,-1){ \small (a) rgb}
        \put(65.5,-1){ \small (b) depth}
        \put(77,-1){ \small (c) normal}
        \put(90,-1){ \small (d) mask}
    \end{overpic}
    \caption{Multi-modal visual understanding on \textbf{ScanNet}. }
    \label{fig_supp:vision_scannet}
\end{figure*}

First, our method relies on pseudo-labels generated by expert models, which can introduce generalization issues. These pseudo-labels may be inaccurate or inconsistent, especially in complex scenes, potentially affecting MMGen’s output quality. Fig. \ref{fig:vis_error_label} illustrates two examples of pseudo-label errors.
For normal estimation, the expert models fail to produce correct outputs for both samples, resulting in missing predictions. For segmentation, in the first sample, the expert model fails to segment the foreground objects, while in the second sample, noticeable boundaries appear between two segmentation regions, as indicated by arrows. Such boundary artifacts are common and are often overfitted by our model (see Fig. \ref{fig_supp:condgen_depth} (e) for an example).
In the future, advancements in expert models could help mitigate these issues. Additionally, exploring novel training strategies may further alleviate the impact of these artifacts and enhance the robustness of our model.

Secondly, compared to large-scale 2D foundation models, our model and dataset size are relatively limited. Expanding the model size and incorporating a larger training dataset, such as extensive synthetic data, could enhance generation quality and diversity. 

We leave addressing these limitations as a direction for future work.



\end{document}